\documentclass[Afour,sageh,times]{sagej}

\usepackage{moreverb,url}

\usepackage[colorlinks,bookmarksopen,bookmarksnumbered,citecolor=red,urlcolor=red]{hyperref}

\newcommand\BibTeX{{\rmfamily B\kern-.05em \textsc{i\kern-.025em b}\kern-.08em
T\kern-.1667em\lower.7ex\hbox{E}\kern-.125emX}}

\def\volumeyear{2024}

\usepackage{enumitem}
\usepackage{booktabs}
\usepackage{caption}
\usepackage{xcolor}
\hypersetup{colorlinks=true}
\usepackage{mathtools}
\usepackage{amsfonts}
\usepackage{mathtools}
\usepackage{lipsum}
\usepackage{multicol}
\usepackage{multirow}
\usepackage{subcaption}
\usepackage{tcolorbox}

\usepackage{tabularx}
\usepackage{colortbl}
\usepackage[normalem]{ulem}

\definecolor{dgreen}{rgb}{0,0,0}
\definecolor{dyellow}{rgb}{.7,.7,0}
\definecolor{dred}{rgb}{1,0,0}
\definecolor{dblue}{rgb}{0,0,0.7}
\definecolor{dorange}{rgb}{0.9,0.5,0.1}
\definecolor{light-gray}{rgb}{0.8, 0.8, 0.8}
\definecolor{highlight}{HTML}{e3eeff}
\definecolor{comment-green}{rgb}{0.435, 0.576, 0.106}
\definecolor{prompt-gray}{HTML}{a7a7a7}
\definecolor{code-syntax}{HTML}{0060b1}
\definecolor{func}{HTML}{d6d69f}
\definecolor{args}{HTML}{90d6fd}

\newcommand{\ie}{i.e., }
\newcommand{\eg}{e.g., }

\newcommand{\command}[1]{\textcolor{comment-green}{#1}}
\newcommand{\prompt}[1]{\textcolor{prompt-gray}{#1}}

\usepackage{inconsolata}
\usepackage[T1]{fontenc}

\newcommand{\hlcode}[1]{\colorbox{highlight}{\makebox[0.96\linewidth][l]{#1}}}

\newcommand{\lmp}[1]{
\begin{tcolorbox}[boxsep=0pt,
                  left=3pt,
                  right=-4pt,
                  top=3pt,
                  bottom=3pt,
                  arc=0pt,
                  boxrule=0.5pt,
                  colframe=light-gray,
                  colback=white
                  ]
\small{  
\ttfamily
#1
}
\end{tcolorbox}
}

\newcommand{\speciallmp}[1]{
\begin{tcolorbox}[
 enlarge top by=0.5em,
 boxsep=0pt,
                  left=3pt,
                  right=-4pt,
                  top=3pt,
                  bottom=3pt,
                  arc=0pt,
                  boxrule=0.5pt,
                  colframe=light-gray,
                  colback=white
                  ]
\small{  
\ttfamily
#1
}
\end{tcolorbox}
}

\begin{document}


\setcounter{page}{1}
\setcounter{section}{0}

\runninghead{Huang~\textit{et~al.}}

\title{Multimodal Spatial Language Maps\\ for Robot Navigation and Manipulation}


\author{Chenguang Huang\affilnum{1}, Oier Mees\affilnum{2}, Andy Zeng\affilnum{3} and Wolfram Burgard\affilnum{1}}

\affiliation{\affilnum{1}University of Technology Nuremberg, Germany\\
\affilnum{2}UC Berkeley, USA\\
\affilnum{3}Google Research, USA}
\corrauth{Chenguang Huang, 
University of Technology Nuremberg,
Department of Computer Science and Artificial Intelligence,
Artificial Intelligence and Robotics Lab,
Ulmenstraße 52i,
90461 Nuremberg, Germany}

\email{chenguang.huang@utn.de}

\begin{abstract}
Grounding language to a navigating agent’s observations can leverage pretrained multimodal foundation models to match perceptions to object or event descriptions. However, previous approaches remain disconnected from environment mapping, lack the spatial precision of geometric maps, or neglect additional modality information beyond vision. To address this, we propose multimodal spatial language maps as a spatial map representation that fuses pretrained multimodal features with a 3D reconstruction of the environment. We build these maps autonomously using standard exploration. We present two instances of our maps, which are visual-language maps (VLMaps) and their extension to audio-visual-language maps (AVLMaps) obtained by adding audio information. When combined with large language models (LLMs), VLMaps can translate natural language commands into open-vocabulary spatial goals (\eg ``in between the sofa and TV'') directly localized in the map, and be shared across different robot embodiments to generate tailored obstacle maps on demand. Building upon the capabilities above, AVLMaps extend VLMaps by introducing a unified 3D spatial representation integrating audio, visual, and language cues through the fusion of features from pretrained multimodal foundation models. This enables robots to ground multimodal goal queries (\eg text, images, or audio snippets) to spatial locations for navigation. Additionally, the incorporation of diverse sensory inputs significantly enhances goal disambiguation in ambiguous environments. Experiments in simulation and real-world settings demonstrate that our multimodal spatial language maps enable zero-shot spatial and multimodal goal navigation and improve recall by 50\% in ambiguous scenarios. These capabilities extend to mobile robots and tabletop manipulators, supporting navigation and interaction guided by visual, audio, and spatial cues. Code and videos are available at \href{https://mslmaps.github.io}{https://mslmaps.github.io}.
\end{abstract}


\maketitle

\section{Introduction}
\label{sec:introduction}

People are excellent navigators of the physical world due in part to their remarkable ability to build cognitive maps~\citep{mcnamara1989subjective} that form the basis of spatial memory~\citep{chun1998contextual,newman2007learning} to (i) localize landmarks at varying ontological levels, such as a book; on the shelf; in the living room, or to (ii) determine whether the layout permits navigation between two points.
Meanwhile, humans exhibit a remarkable ability to integrate and leverage multiple sensing modalities to efficiently move around in the physical world. Our actions are driven by a myriad of sensory cues: the sound of glass breaking might signal a dangerous situation, the microwave might buzz to indicate it is done, or a dog might bark to draw our attention. Acoustic signals particularly represent a valuable complementary form of information, also evident by the utility that they provide for the visually impaired, who may rely on them for navigation. Research in cognitive science also suggests that children understand and integrate information from different sensing modalities into spatial cognitive maps~\citep{kording2007causal}.

Classic methods for robot navigation~\citep{thrun1998probabilistic, endres2012evaluation} build geometric maps for path planning. Although some previous extensions parse goals from templated natural language commands~\citep{tellex2011understanding, macmahon2006walk}, they struggle to generalize to unseen instructions. Learning methods directly optimize for navigation policies grounded in language end-to-end (commands to actions)~\citep{anderson2018vision, anderson2021sim} but require copious amounts of data. Recent works demonstrate that multimodal foundation models~\citep{radford2021learning,li2021language} pretrained on Internet-scale data (\eg images and their captions) can be used out-of-the-box to ground language to the visual observations of a navigating agent, without additional data collection or model fine-tuning. These models enable mobile robots to handle new instructions that specify unseen object goals and can be combined with exploration algorithms to search for the first instance of any object (CoW)~\citep{gadre2022clip} or traverse object-centric landmarks in graphs (LM-Nav)~\citep{shah2022lm}. While promising, these methods predominantly use vision language models (VLMs) as critics to match image observations to object goal descriptions, but remain disjoint from the mapping of the environment, lacking the fine-grained spatial precision of classic geometric maps. Furthermore, they neglect the great potential of information in other sensing modalities such as audio. Therefore, these methods struggle to (i) localize spatial goals \eg ``in between the sofa and the TV'', to (ii) build persistent representations that can be shared across different embodiments, \eg mobile robots, drones, or to (iii) localize multimodal goals \eg ``the sound of the baby crying'', or an image of a refrigerator. How to best spatially anchor various sensing modalities, including visual and audio signals, in ways that enable effective data-efficient cross-modal reasoning for downstream robotics tasks, remains a relatively open question.

In this work, we address that question by introducing Multimodal Spatial Language Maps, a general mapping framework that is (i) spatial, (ii) multimodal, (iii) reusable across different robot embodiments, and (iv) readily extensible to additional sensing modalities in the future. At the heart of our approach are two concrete map instances: Visual-Language Maps (VLMaps) and their multimodal extension, Audio-Visual-Language Maps (AVLMaps). We begin by evaluating VLMaps, which fuse pretrained visual-language features from image observations with a 3D reconstruction of the environment. This fusion makes VLMaps both spatial-preserving, enabling localizing queries like ``in between the sofa and the TV'' in the map, and reusable across embodiments, since the same voxelized map can generate tailored obstacle grids for different robots by defining different sets of obstacle categories. VLMaps can be built from a robot’s video stream using standard exploration strategies. When coupled with a large language model (LLMs) in a Socratic fashion~\citep{zeng2022socratic}, they translate long-horizon natural-language commands into sequences of open-vocabulary, spatially grounded goals.

 \begin{figure}[t]
	\centering
	\includegraphics[width=1\columnwidth]{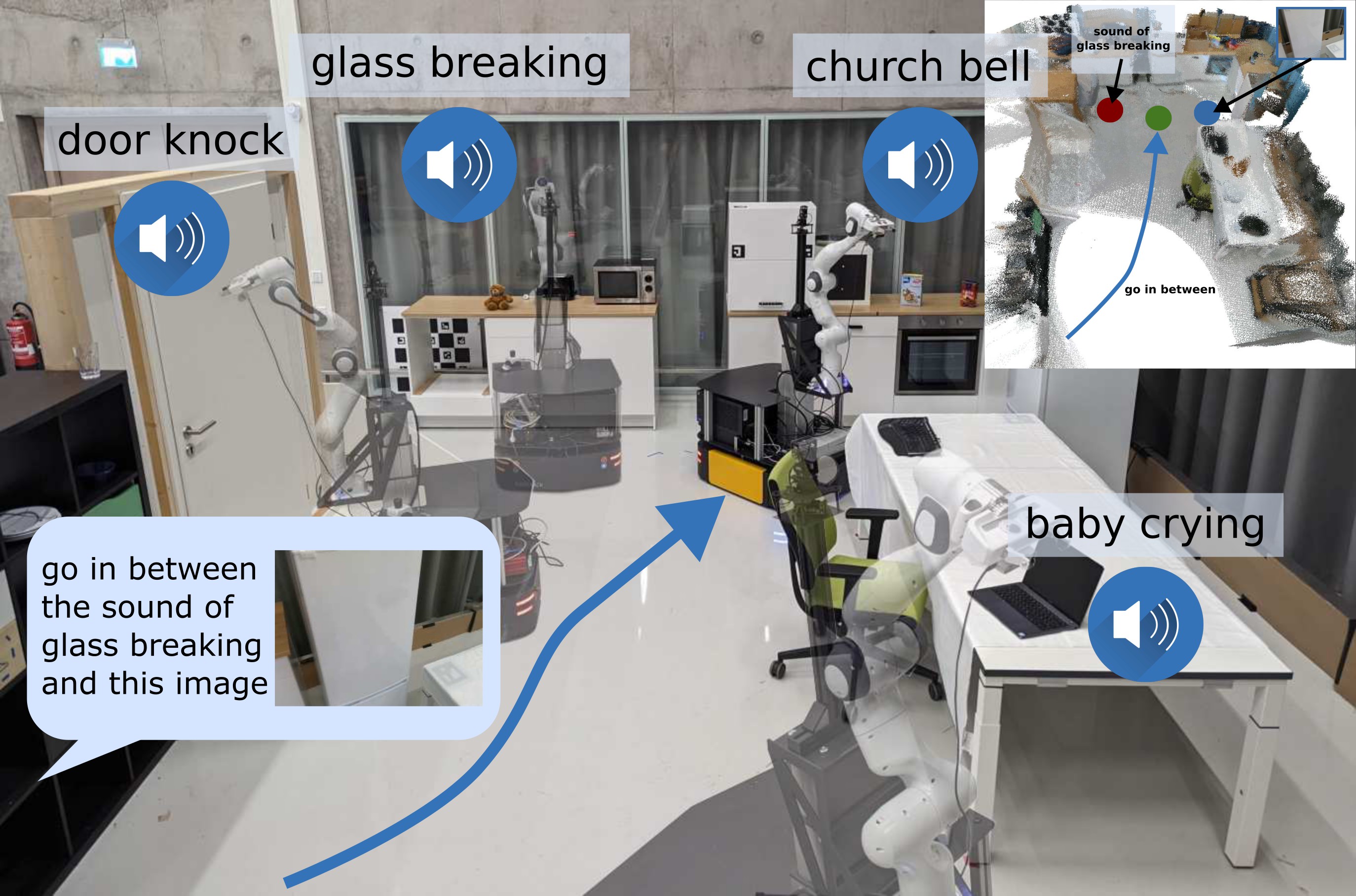}
	\caption{\small \textbf{AVLMaps} provide an open-vocabulary 3D map representation for storing cross-modal information
from audio, visual, and language cues. When combined with large language models, AVLMaps 
consumes multimodal prompts from audio, vision, and language to solve zero-shot spatial goal navigation by effectively leveraging complementary information sources to disambiguate goals.}
	\label{fig:cover_lady}

\end{figure}

Subsequently, we extend VLMaps to Audio-Visual-Language Maps, AVLMaps, a unified 3D spatial map representation for storing cross-sensing information from audio, visual, and language modalities. By introducing a new modality, audio, we extrapolate the capabilities of our framework to the multimodal setting, showing its extensibility to additional sensing modalities. AVLMaps can be built from image and audio observations captured during reconstruction, by computing dense pre-trained features from open-vocabulary multimodal foundation models trained on Internet-scale data~\citep{li2022languagedriven,ghiasi2021open,guzhov2022audioclip} and fusing them into a shared 3D voxel grid representation. Beyond VLMaps, AVLMaps:
\begin{itemize}[leftmargin=*]
  \item allow for landmarks (or areas and regions of interest) indexing in the environment via open-vocabulary \emph{multimodal queries} (\eg abstract textual descriptions, images, or audio snippets), enabling downstream applications including multimodal goal-driven navigation, without domain-specific model finetuning.
  \item include audio information, which allows robots to more often correctly disambiguate goal locations using sound (\eg ``go to the table where you heard coughing'' in environments where there are multiple tables, etc). 
  \item extend the spatial characteristic of VLMaps to the multimodal domain, enabling zero-shot \emph{multimodal} \emph{spatial} goal localization, \eg ``Go in between the \{image of a refrigerator\} and the sound of breaking glass'' as in Fig.~\ref{fig:cover_lady}.
\end{itemize}

\noindent{}Extensive experiments in both simulated and real-world settings demonstrate that our VLMaps enable more effective long-horizon language-conditioned spatial goal navigation than baseline alternatives, such as CoW~\citep{gadre2022clip} and LM-Nav~\citep{shah2022lm}. They can be shared across different robot embodiments to generate tailored obstacle maps for efficient, embodiment-specific path planning. Building on this, AVLMaps further extend these capabilities to navigating to goal locations specified by, \eg natural language descriptions of sounds or visual landmarks -- and notably, can disambiguate multiple possible goal locations using multimodal information, (using object semantics to pinpoint one of the multiple possible sound goals, or using vision to pinpoint one of the multiple possible locations where similar objects were found) quantitatively better than unimodal baseline alternatives by up to 50\% in top-1 recall in ambiguous scenarios. This article expands our previous work~\citep{huang23vlmaps,huang23avlmaps} by expanding our evaluation to demonstrate that AVLMaps' capabilities continue to naturally improve with better-performing pre-trained audio-language foundation models such as AudioCLIP~\citep{guzhov2022audioclip} and CLAP~\citep{elizalde2023clap} and that the achieved multimodal disambiguation capabilities also translate to challenging robot manipulation tasks. AVLMaps are simple and effective in leveraging multiple multimodal foundation models together in tandem to reach broader language-driven robot navigation capabilities, but are also not without limitations -- we discuss these and avenues for future work. The code is available at \href{https://mslmaps.github.io/}{https://mslmaps.github.io/}.
\section{Related Work}
\label{sec:related_work}
\textbf{Semantic Mapping.}
 In recent years, the synergy of the traditional SLAM techniques and the advancements in vision-based semantic understanding has led to augmenting 3D maps with semantic information~\citep{salas2013slam++, mccormac2017semanticfusion}. Stemming from the intuition of augmenting 3D points in the map with 2D segmentation results, previous works focus on either abstracting the map at object-level with a pose graph~\citep{mccormac2018fusion++} or an octree~\citep{xu2019mid} or modeling the dynamics of objects in the map~\citep{runz2018maskfusion}. Despite lifting the 3D reconstruction to a semantic level, these methods are restricted to a predefined set of semantic classes. Recent works like LM-Nav~\citep{shah2022lm}, CoW~\citep{gadre2022clip}, VLMaps~\citep{huang23vlmaps}, NLMap-SayCan~\citep{chen2022open}, OpenScene~\citep{peng2022openscene}, or CLIP-Fields~\citep{shafiullah2022clipfields} have shown that integrating visual-language features, generated by either pre-trained or fine-tuned models, into a topological graph or an occupancy map enables open-vocabulary object indexing with natural language, freeing the maps from fixed-size semantic categories. Recent approaches also investigate other open-vocabulary map representations such as NeRF~\citep{engelmann2024opennerf,kerr2023lerf,kim2024garfield}, Gaussian Splattings~\citep{qin2024langsplat,zuo2024fmgs}, and Scene Graphs~\citep{gu2024conceptgraphs,werby2024hovsg}. However, these works focus on visual perception to map and move through an environment, overlooking complementary sources of information such as acoustic signals. In contrast, AVLMaps integrate audio, visual, and language cues into a 3D map, equipping the agent with the ability to navigate to multiple types of multimodal goals and effectively disambiguate goals.

\noindent\textbf{Vision and Language Navigation.}
Recently, also Vision-and-Language Navigation (VLN) has received increased attention~\citep{anderson2018vision, krantz2020beyond}. Further work has focused on learning end-to-end policies that can follow route-based instructions on topological graphs of simulated environments~\citep{anderson2018vision, fried2018speaker, guhur2021airbert}. However, agents trained in this setting do not have low-level planning capabilities and rely heavily on the topological graph, limiting their real-world applicability~\citep{anderson2021sim}. Moreover, despite extensions to continuous state spaces~\citep{krantz2020beyond, krantz2021waypoint, hong2022bridging}, most of these learning-based methods are data-intensive. The recent success of large pretrained vision and language models~\citep{radford2021learning,brown2020language} has spurred a flurry of interest in applying their zero-sot capabilities to open-vocabulary object navigation~\citep{shah2022lm,gadre2022clip}. LM-Nav~\citep{shah2022lm} combines three pre-trained models to navigate via a topological graph in the real world. CoW~\citep{gadre2022clip} performs zero-shot language-based object navigation by combining CLIP-based~\citep{radford2021learning} saliency maps and traditional exploration methods. However, both methods are limited to navigating to object landmarks and are less capable of understanding finer-grained queries, such as ``to the left of the chair'' and ``in between the TV and the sofa''. In contrast, our method, VLMaps, enables spatial language indexing beyond object-centric goals and can generate open-vocabulary obstacle maps. Our extension, AVLMaps, further enables multimodal spatial concept indexing such as ``between the {image of a
refrigerator} and the sound of breaking glass''.

\noindent\textbf{Multimodal Navigation.}
Recent advances in simulation applications~\citep{habitat19iccv,kolve2017ai2,chen2020soundspaces,gan2021threedworld} have boosted research on multimodal navigation in two distinct directions: (i) vision-and-language navigation (VLN)~\citep{anderson2018vision,krantz2020beyond} where an agent needs to follow a natural language instruction towards the goal with visual input, and (ii) audio-visual navigation (AVN)~\citep{chen2020soundspaces} in which an agent should navigate to the sound source based on information from a binaural sensor and vision. Despite different degrees of success in both directions~\citep{fried2018speaker,guhur2021airbert,chen2020learning,younes2023catch,gan2020look,gan2022finding}, less attention has been paid to solving the navigation problem involving vision, language, and audio at the same time. The most relevant concept to our knowledge is from AVLEN~\citep{paul2022avlen}, which extends the AVN with a further query step, introducing a language instruction that helps with navigating to the sound source. In addition, most of the existing methods on AVN focus on approaching the sound without understanding its semantics. In our work, we propose a method to integrate both visual and sound semantics into the same map, enabling a robot to navigate to multimodal goals specified with either goal image or natural language like ``go to the sound of baby crying'', ``go to the table'' or multimodal prompts such as ``go to the \{image of a table\} where the sound of the microwave was heard''.

 \begin{figure*}[t]
	\centering
	\includegraphics[width=1\textwidth]{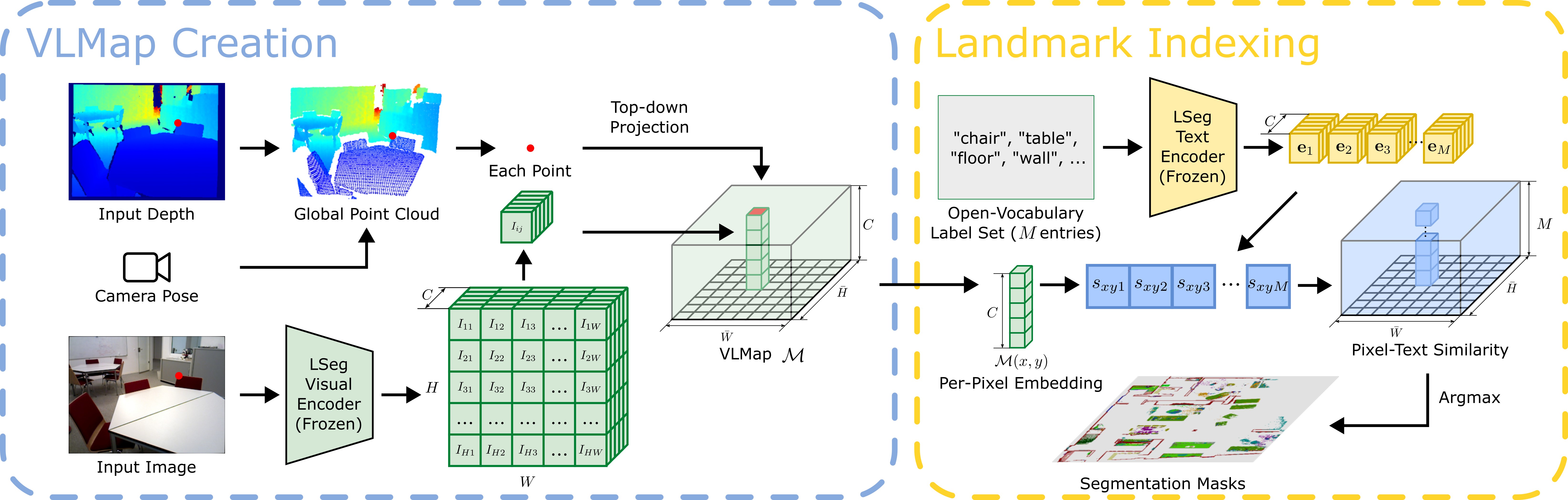}
	\caption{\small The creation and language-conditioned indexing of a VLMap. A VLMap is created by fusing pretrained visual-language features into the reconstruction of the environment to enable visual-spatial-language-based reasoning. By providing a list of open-vocabulary labels, we retrieve segmentation masks for semantic classes required by downstream applications.}
	\label{fig:vlmaps_pipeline}

\end{figure*}

\noindent\textbf{Pre-trained Zero-shot Models in Robotics.}
Recent trends have shown that pre-trained foundation models~\citep{radford2021learning,brown2020language,liang2023ovseg} serve as powerful tools for robotic tasks. Most works exploit the inherent perception and reasoning abilities of Vision Language Models (VLMs) or Large Language Models (LLMs) trained with cloud-sourced data to boost the performance of robot tasks including object detection and segmentation~\citep{kamath2021mdetr,gu2021open,li2021language}, robot manipulation~\citep{shridhar2022cliport,liang2022code,mees2022calvin,mees2022hulc,mees22hulc2,rosete2022tacorl,Zawalski24-ecot,chen2024vision,zhou2024autonomous}, and navigation~\citep{shah2022lm,gadre2022clip,chen2022nlmapsaycan,huang23vlmaps,hirose24lelan,gu2024conceptgraphs,werby2024hovsg}. With access to a growing volume of robot control data annotated with semantic language labels, recent approaches have advanced the training of a Vision-Language-Action (VLA) model, bridging the gap between visual-language comprehension and low-level robot control within a unified foundational model~\citep{brohan2022rt1,zitkovich2023rt2,o2024rtx,kim24openvla,octo_2023,Doshi24-crossformer}.
Despite promising results from previous methods, little effort has been made to exploit the audio-language pre-trained models (ALMs)~\citep{guzhov2022audioclip,elizalde2023clap} in robotic tasks. In this work, we leverage the foundation models focusing on different modalities, e.g. audio, language, and vision, to create a mapping pipeline to understand multimodal information in the scene and achieve robot navigation given language, image, or audio description queries. Concurrent work ConceptFusion~\citep{jatavallabhula2023conceptfusion} demonstrates that audio can be used as queries to index locations in a visual-language map. However, it doesn't support integrating audio information from the observation data into the representation of the scene as we do in this work.

\section{Method}
\label{sec:method}

Our goals are two-fold: (i) build a visual language map representation that synergizes the open-vocabulary capability of vision-language models and the spatial characteristic of geometric maps, and (ii) extend such a map to a multimodal spatial language map representation, in which object landmarks (``sofa''), areas (``kitchen''), audio semantics (``the sound of a baby crying''), or visual goals can be directly localized using natural language or a target image. To achieve the first goal, we propose VLMaps, which can be constructed with pre-trained visual-language models by consuming RGB-D streaming data and camera odometry. Such a map allows for spatial goal indexing like ``to the left of the sofa'', and can dynamically adapt to different embodiments, enabling users to freely define obstacle categories for the robot to generate a customized occupancy map for path planning. For the second goal, we propose a more general representation, AVLMaps, which takes VLMaps as one sub-module and extends it to consume multimodal data during mapping and support multimodal concept indexing, such as audio, images, and language. We also propose a cross-modal reasoning method to disambiguate locations referring to targets from different modalities (``the sound of brushing teeth near the sink'', or ``the table near this image: \{image\}''). In the following subsections, we first start with (i) how to build a VLMap by integrating visual-language features into spatial map location (Sec.~\ref{sec:sub_map_creation}), (ii) how to use this map to localize open-vocabulary landmarks (Sec.~\ref{sec:sub_landmark_indexing}), (iii) how we can build open-vocabulary obstacle maps from a list of obstacle categories for different robot embodiments (Sec.~\ref{sec:obstacle_maps}), and (iv) how we can use this map for spatial goal navigation with the help of an LLM (Sec.~\ref{sec:sub_language_spatial_navigation}). Later, we consider VLMaps in a broader context, and demonstrate (v) how to extend it to a multimodal map that integrates audio, language, and visual information in the map and use it for localizing different targets (Sec.~\ref{subsec_building_avlmaps}), (vi) how to disambiguate goal locations with multimodal information (Sec.~\ref{subsec_cross_modal_reasoning}), and (vii) how such a multimodal map can be used with large language models (LLMs) for multimodal goal navigation, without additional data collection or model fine-tuning (Sec.~\ref{subsec_multimodal_nav_from_language}). We show the pipeline of building a VLMap in Fig.~\ref{fig:vlmaps_pipeline}, and later show the system pipeline of our multimodal map representation in Fig.~\ref{fig:system_overview}.

\subsection{Building a Visual-Language Map}
\label{sec:sub_map_creation}
The key idea behind VLMaps is to fuse pretrained visual-language features with a 3D reconstruction. 
We achieve this by computing dense pixel-level embeddings from an existing visual-language model (over the video feed of the robot) and by back-projecting them onto the 3D surface of the environment (captured from depth data used for reconstruction with visual odometry). The overview of VLMaps creation is shown on the left of Fig.~\ref{fig:vlmaps_pipeline}.

In our work, we utilize LSeg~\citep{li2021language} as the visual-language model, a language-driven semantic segmentation model that segments the RGB images based on a set of free-form language categories. The LSeg visual encoder maps an image such that the embedding of each pixel lies in the CLIP feature space. In our approach, we fuse the LSeg pixel embeddings with their corresponding 3D map locations. In this way, without explicit manual segmentation labels, we incorporate a powerful language-driven semantic prior that inherits the generalization capabilities of VLMs. The only assumption we make is access to odometry, which is readily available from RGB-D SLAM systems and enables us to build a map from sequences of RGB-D images.

Formally, we define VLMap as $\mathcal{M} \in \mathbb{R}^{\bar{H} \times \bar{W} \times C}$, where $\bar{H}$ and $\bar{W}$ represent the size of the top-down grid map, and $C$ represents the length of the VLM embedding vector for each grid cell. Together with the scale parameter $s$ (meters per pixel), a VLMap $\mathcal{M}$ represents an area with size $s\bar{H} \ \text{meters} \times s\bar{W} \ \text{meters}$.
To build the map, for each RGB-D frame, we back-project all the depth pixels $\mathbf{u} = (u, v)$ to form a local depth point cloud that we transform to the world frame, $\mathbf{P}_{k} = D(\mathbf{u}) K^{-1} \mathbf{\tilde{u}}$ and $\mathbf{P}_{W} = T_{Wk} \mathbf{P}_{k}$ where $\mathbf{\tilde{u}} = (u, v, 1)$, $K \in \mathbb{R}^{3\times 3}$ is the intrinsic matrix of the depth camera, $D(\mathbf{u}) \in \mathbb{R}$ is the depth value of the pixel $\mathbf{u}$, $T_{Wk}$ is the transformation from the world coordinate frame to the k-th camera frame, $\mathbf{P}_{k} \in \mathbb{R}^{3}$ is the 3D point position in the k-th frame, and $\mathbf{P}_{W} \in \mathbb{R}^{3}$ is the 3D point position in the world coordinate frame. We then project the point $\mathbf{P}_{W}$ to the ground plane and get the pixel $\mathbf{u}$'s corresponding position on the grid map,
\begin{equation}
\label{eq:px_map}
p^{x}_{map} = \Bigl\lfloor \frac{\bar{H}}{2} + \frac{P^{x}_{W}}{s} + 0.5 \Bigr\rfloor,~ p^{y}_{map} = \Bigl\lfloor \frac{\bar{W}}{2} - \frac{P^{z}_{W}}{s} +0.5 \Bigr\rfloor
\end{equation}
where $p^{x}_{map}$ and $p^{y}_{map}$ represent the coordinates of the projected point in the map $\mathcal{M}$.

Once we build the grid map, we apply LSeg's visual encoder $f(\mathcal{I}): \mathbb{R}^{H \times W \times 3} \rightarrow \mathbb{R}^{H \times W \times C}$ to the RGB image $\mathcal{I}_{k}$ and generate the pixel-level embedding $\mathcal{F}_{k} \in \mathbb{R}^{H \times W \times C}$. Given the RGB-D registration, we project each image pixel $\mathbf{u}$'s embedding $\mathbf{q} = \mathcal{F}_{k}(\mathbf{u}) \in \mathbb{R}^{C}$ to its corresponding grid cell location $(p^{x}_{map}, p^{y}_{map})$ in the top-down grid map.  Intuitively, there exist multiple 3D points projecting to the same grid location in the map. Thus, we average their embeddings, $\mathcal{M} ( p^{x}_{map},\ p^{y}_{map} ) = \frac{1}{n}  \sum_{i=1}^{n} \mathbf{q}_i$
where $\mathcal{M} ( p^{x}_{map},\ p^{y}_{map} ) \in \mathbb{R}^{C}$ represents the map features at the grid position $(p^{x}_{map}, p^{y}_{map})$, $n$ represents the total number of points projecting to the grid location $(p^{x}_{map},\ p^{y}_{map})$, and $\mathbf{q}_i \in \mathbb{R}^{C}$ denotes the corresponding pixel embedding of each point. We note that these $n$ points might not only come from a single frame, but also from points from multiple frames. Therefore, the resulting features contain the averaged embeddings from multiple views of the same object.

\subsection{Localizing Open-Vocabulary Landmarks}
\label{sec:sub_landmark_indexing}
We now describe how to localize landmarks in VLMaps with free-form natural language. The overview of the indexing process is shown on the right of Fig.~\ref{fig:vlmaps_pipeline}.
Formally, we define the input language list as $\mathcal{L} = [\mathbf{l}_0, \mathbf{l}_1, \ldots, \mathbf{l}_M]$ where $\mathbf{l}_i$ represents the i-th category in text form, and $M$ represents the number of categories defined by the user. Some examples of the input language list are [``chair'', ``sofa'', ``table'', ``other''] or [``furniture'', ``floor'', ``other'']. As Li \emph{et al.} \citep{li2021language}, we apply the pre-trained CLIP text encoder \citep{radford2021learning} to convert such list of texts into a list of vector embeddings $[\mathbf{e}_0, \mathbf{e}_1, \ldots, \mathbf{e}_M],\ \mathbf{e} \in \mathbb{R}^{C}$, which are organized into an embedding matrix $E \in \mathbb{R}^{M \times C}$, where each row of the matrix represents the embedding of a category. The map embeddings $\mathcal{M}$ are also flattened into a matrix $Q \in \mathbb{R}^{\bar{H}\bar{W} \times C}$, where each row represents the embedding of a pixel in the top-down grid map. We then compute the pixel-to-category similarity matrix $S = Q \cdot E^{T}$,
where $S \in \mathbb{R}^{\bar{H}\bar{W} \times M}$. Each element $S_{ij}$ in the matrix stores the similarity value between a pixel and a text category, indicating how likely this pixel belongs to the class. By applying the $\mathrm{argmax}$ operator along the row direction to $S$ and reshaping the resulting vector to shape $\bar{H} \times \bar{W}$, we get the final segmentation result $R \in \mathbb{R}^{\bar{H} \times \bar{W}}$. Each element $R_{ij}$ represents the label index of the input language list $\mathcal{L}$ at the grid map location $(i, j)$. With the final resulting matrix $R$, we compute the most related language-based category for every pixel in the grid map.

\begin{figure}[t]
	\centering
	\includegraphics[width=1\columnwidth]{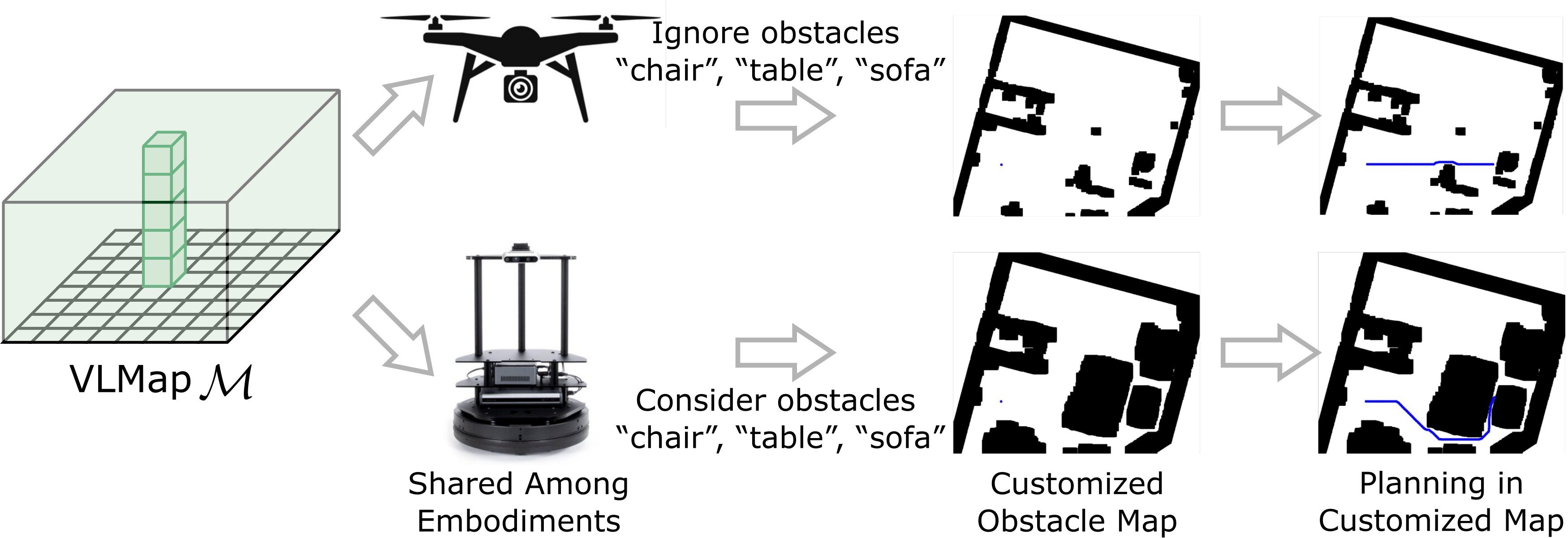}
	\caption{\small The overview of building customized obstacle maps for different robot embodiments. By specifying different obstacle categories in natural language for different embodiments, different obstacle maps can be built to ensure the most efficient path planning for different embodiments.}
	\label{fig:cross_embodiment_obstacle_map}

\end{figure}

\subsection{Generating Open-Vocabulary Obstacle Maps}
\label{sec:obstacle_maps}
Building a VLMap enables us to generate obstacle maps that inherit the open-vocabulary nature of the VLMs used (LSeg and CLIP). Specifically, given a list of obstacle categories described with natural language, we can localize those obstacles at runtime to generate a binary map for collision avoidance and/or shortest path planning, as is shown in Fig.~\ref{fig:cross_embodiment_obstacle_map}. A prominent use case for this is sharing a VLMap of the same environment between different robots with different embodiments (\ie cross-embodiment problem \citep{zakka2022xirl,ganapathi2022implicit}), which may be useful for multi-agent coordination \citep{wu2021spatial}. For example, a large mobile robot may need to navigate around a table (or other large furniture), while a drone can directly fly over it. By simply providing two different lists of obstacle categories -- one for the large mobile robot (that contains ``table''), and another for the drone (that does not), we can generate two distinct obstacles maps for the two robots to use respectively, sourced on-the-fly from the same VLMap.

To do so, we first extract an obstacle map $\mathcal{O} \in \{0, 1\}^{\bar{H} \times \bar{W}}$ where each projected position of the depth point cloud in the top-down map is assigned~1, and otherwise~0. To avoid points from the floor or the ceiling, points $P_{W}$ are filtered out depending on their height,
\begin{equation}
\mathcal{O}_{ij} = 
    \begin{dcases}
        1,& t_1 \leq P^{y}_{W} \leq t_2\ and\ p^{x}_{map} = i\ and\ p^{y}_{map} = j\\
        0,& \text{otherwise}
    \end{dcases}
\end{equation}
where $t_1, t_2 \in \mathbb{R}$ are the lower and upper thresholds for the $y$-component (we define $y$ axis to be along the height direction) of the point $P_{W}$. 
Second, to obtain obstacle maps tailored to a certain embodiment, we define a list of potential obstacle categories $\mathcal{L}_{obs} = [\mathbf{l}_{obs0}, \mathbf{l}_{obs1}, \ldots, \mathbf{l}_{obsM}]$, where $\mathbf{l}_{obsi}$ represents the i-th obstacle category in language, and $M$ represents the total number of obstacle categories defined by the user. We then apply the open-vocabulary landmark indexing introduced in Sec.~\ref{sec:sub_landmark_indexing} and obtain segmentation masks for all defined obstacles. For a specific embodiment $k$, we choose a subset of classes out of the whole potential obstacle list $\mathcal{L}_{obs}$ and take the union of their segmentation masks to get the obstacles mask $\tilde{\mathcal{O}}_{em_k}$. We ignore false predictions of obstacles on floor region in $\tilde{\mathcal{O}}_{em_k}$ by taking the intersection with $\mathcal{O}$ to get the final obstacle map $\mathcal{O}_{em_k}$.

\subsection{Zero-Shot Spatial Goal Navigation from Language}
\label{sec:sub_language_spatial_navigation}
In this section, we describe our approach to long-horizon (spatial) goal navigation, given a set of landmark descriptions specified by natural language instructions such as
\lmp{
\prompt{
move first to the left side of the counter, then \\
move between the sink and the oven, then move back \\
and forth to the sofa and the table twice
}
}
\noindent Notably different from prior work~\citep{gadre2022clip, shah2022lm}, VLMaps allow us to reference precise spatial goals such as: ``in between the sofa at the TV'' or ``three meters to the east of the chair.''
Specifically, we use a large language model (LLM) to interpret the input natural language commands and break them down into subgoals~\citep{ahn2022can,shah2022lm,zeng2022socratic}. In contrast to prior work, which may reference these subgoals with language and map to low-level policies with semantic translation~\citep{huang2022language} or affordances~\citep{ahn2022can,huang2022inner,zeng2019learning}, we leverage the code-writing capabilities of LLMs to generate executable Python robot code~\citep{liang2022code,mees22hulc2,chen2021evaluating,brown2020language} that can (i) make precise calls to parameterized navigation primitives, and (ii) perform arithmetic when needed. The generated code can directly be executed on the robot with the built-in Python \textbf{exec} function.

Note that recent works~\citep{liang2022code,mees22hulc2,chen2021evaluating,brown2020language} have shown that code-writing language models (\eg{} Codex~\citep{chen2021evaluating}) trained on billions of lines of code from Github can be used to synthesize new simple Python programs from docstrings. In this work, we re-purpose these models for mobile robot planning by priming them with several input examples of natural language commands (formatted as comments) paired with corresponding robot code (via few-shot prompting). The robot code can express functions or logic structures (if-then-else statements or for/while loops) and parameterize API calls (\eg{} \texttt{move\_to(target\_name)} or \texttt{turn(degrees)}. The full list is available in Table~\ref{table:spatial_nav_primitives}) that map to spatial behaviors specified by the language commands. The full prompt is shown in Fig.~\ref{fig:spatial_goal_full_prompt}.

\begin{figure}[htp]
    \centering
    \lmp{
    \prompt{
    \# move a bit to the right of the fridge\\
    robot.move\_to\_right(`refrigerator')\\
    \\
    \# move in between the couch and bookshelf\\
    robot.move\_in\_between(`couch', `bookshelf')\\
    \\
    \# face the toilet\\
    robot.face(`toilet')\\
    \\
    \# move to the west of the chair\\
    robot.move\_west(`chair')\\
    \\
    \# turn right 20 degrees\\
    robot.turn(20)\\
    \\
    \# find any chairs in the environment\\
    robot.move\_to\_object(`chair')\\
    \\
    \# with the television on your left\\
    robot.with\_object\_on\_left(`television')\\
    \\
    \# move forward for 3 meters\\
    robot.move\_forward(3)\\
    \\
    \# move right 2 meters\\
    robot.turn(90)\\
    robot.move\_forward(2)\\
    \\
    \# move back and forth to the chair and table \\
    \# 3 times\\
    pos1 = robot.get\_pos(`chair')\\
    pos2 = robot.get\_pos(`table')\\
    for i in range(3):\\
    \hspace*{4mm}robot.move\_to(pos1)\\
    \hspace*{4mm}robot.move\_to(pos2)\\
    \\
    \# move 3 meters south of the chair\\
    robot.move\_south(`chair')\\
    robot.face('chair')\\
    robot.turn(180)\\
    robot.move\_forward(3)\\
    \\
    \# turn west\\
    robot.turn\_absolute(-90)\\
    \\
    \# turn east\\
    robot.turn\_absolute(90)\\
    \\
    \# turn south\\
    robot.turn\_absolute(180)\\
    \\
    \# turn north\\
    robot.turn\_absolute(0)\\
    \\
    \# turn east and then turn left 90 degrees \\
    robot.turn\_absolute(90)\\
    robot.turn(-90)\\
    \\
    \# navigate to 3 meters right of the table\\
    robot.move\_to\_right('table')\\
    robot.face('table')\\
    robot.turn(180)\\
    robot.move\_forward(3)\\
    }
    }
    \caption{The full context prompt (prompt in {\color{prompt-gray}gray}) VLMap used for achieving spatial goal navigation tasks in the experiments.}
    \label{fig:spatial_goal_full_prompt}
\end{figure}

\begin{figure*}[!t]
	\centering
	\includegraphics[width=\textwidth]{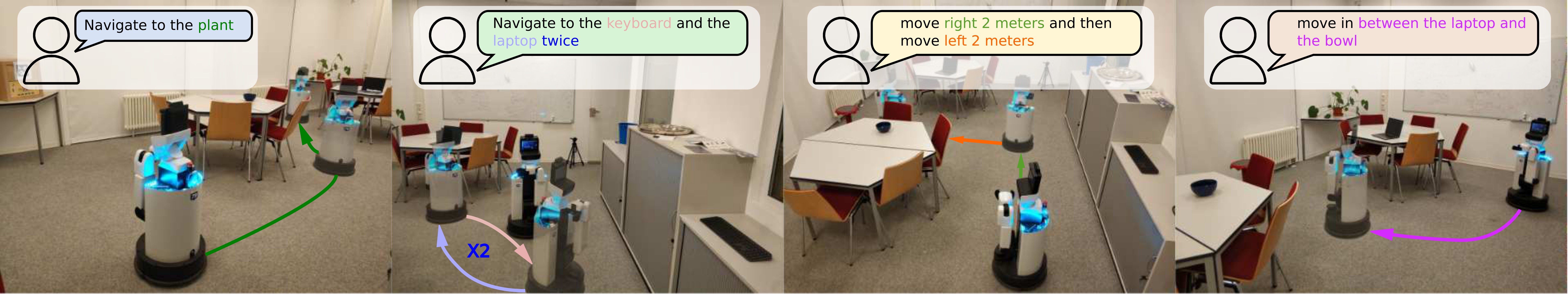}
	\caption{\small VLMaps enable a robot to perform complex zero-shot spatial goal navigation tasks given natural language commands, without additional data collection or model finetuning. }
	\label{fig:spatial_goal_nav}
\end{figure*}

\begin{table*}[!htp]
  \centering
  \begin{tabularx}{\textwidth}{l|X}
  \toprule

    APIs          & Functions                       \\
\midrule

\texttt{\textbf{move\_to}(pos)}         & move to a position on the map.    \\
\rowcolor{highlight}
\texttt{\textbf{move\_to\_left}(object\_name)}         & move to the left side of the nearest front object.    \\
\texttt{\textbf{move\_to\_right}(object\_name)}         & move to the right side of the nearest front object.    \\
\rowcolor{highlight}
\texttt{\textbf{with\_pos\_on\_left}(object\_name)}  & turn until the object is on the robot's left side.     \\
\texttt{\textbf{with\_pos\_on\_right}(object\_name)}  & turn until the object is on the robot's right side.  \\
\rowcolor{highlight}
\texttt{\textbf{move\_in\_between}(object\_a, object\_b)} & move in between two objects. \\
\texttt{\textbf{face}(object\_name)}           & turn until the robot's front is pointing to the object. \\
\rowcolor{highlight}
\texttt{\textbf{turn}(angle)}           & turn right a certain angle. If the angle value is negative, turn left.  \\
\texttt{\textbf{turn\_absolute}(angle)}           & turn to absolute angle. 0 is north, 90 is east, -90 is west, 180 is south. \\
\rowcolor{highlight}
\texttt{\textbf{move\_north}(object\_name)}       & move to the north side of the object. \\
\texttt{\textbf{move\_south}(object\_name)}       & move to the south side of the object. \\
\rowcolor{highlight}
\texttt{\textbf{move\_east}(object\_name)}       & move to the east side of the object. \\
\texttt{\textbf{move\_west}(object\_name)}       & move to the west side of the object. \\
\rowcolor{highlight}
\texttt{\textbf{move\_forward}(dist)}       & move forward ``dist'' meters. \\
  \bottomrule
  \end{tabularx}
  \caption{Navigation API library for spatial goal navigation in VLMaps.}
  \label{table:spatial_nav_primitives}
\end{table*}

\begin{figure}[h]
    \centering
    \speciallmp{
    \command{\# move first to the left side of the counter, then \\
    move between the sink and the oven, then move back \\
    and forth to the sofa and the table twice}\\
    \hlcode{robot.move\_to\_left(`counter')}\\
    \hlcode{robot.move\_in\_between(`sink', `oven')}\\
    \hlcode{pos1 = robot.get\_pos(`sofa')}\\
    \hlcode{pos2 = robot.get\_pos(`table')}\\
    \hlcode{for i in range(2):}\\
    \hlcode{\hspace*{4mm}robot.move\_to(pos1)}\\
    \hlcode{\hspace*{4mm}robot.move\_to(pos2)}\\
    \command{\# move 2 meters north of the laptop, then move 3 \\
    meters rightward}\\
    \hlcode{robot.move\_north(`laptop')}
    \hlcode{robot.face(`laptop')}
    \hlcode{robot.turn(180)}
    \hlcode{robot.move\_forward(2)}
    \hlcode{robot.turn(90)}
    \hlcode{robot.move\_forward(3)}
    }
    \caption{The query and the generated results from the LLM for spatial goal navigation tasks. During the query, the context prompt in Fig.~\ref{fig:spatial_goal_full_prompt} and the input task commands are prompted to the LLM together. The input task commands are in \command{green} and generated outputs are \colorbox{highlight}{highlighted}}
    \label{fig:spatial_goal_query_prompt}
\end{figure}

Spatial goals are defined as positions around the reference object based on spatial descriptions. For example, ``in the middle of the counter and the fridge'', or ``to the left of the sofa'' etc. Traditional object goal navigation methods either directly retrieve the target object's location on the map and plan to it~\citep{shah2022lm,gadre2022clip} or are trained to approach objects as a reactive system~\citep{chaplot2020object_goal_nav}. These methods fall short in reaching spatial goals since these goal locations are free space rather than retrievable locations on semantic maps. However, when we know the reference position, those spatial locations can be computed with simple offsets. The navigation primitive functions (APIs) being called by the language model (\eg{}\texttt{move\_to\_left(`counter')}) use a pre-generated VLMap to localize the coordinates of the open-vocabulary landmarks (``counter'') in the maps (described in Sec.~\ref{sec:sub_landmark_indexing}) modified with predefined scripted offsets (to define ``left''). We then navigate to these coordinates using an off-the-shelf navigation stack~\citep{quigley2009ros} that takes as input the embodiment-specific obstacle map (generated using the same VLMap, with the process described in Sec.~\ref{sec:obstacle_maps}). Some examples of spatial goal navigation in the real world is shown in Fig.~\ref{fig:spatial_goal_nav}.

At test time, the LLM is prompted with context examples (in {\color{prompt-gray}gray}) in Fig.~\ref{fig:spatial_goal_full_prompt} as well as commands (in \command{green}) in Fig.~\ref{fig:spatial_goal_query_prompt}. It can autonomously re-compose API calls to generate new robot code that not only references the new landmarks mentioned in the language commands (as comments), but also can chain together new sequences of API calls to follow unseen instructions accordingly. The inference process is shown in Fig.~\ref{fig:spatial_goal_query_prompt}, and generated outputs are \colorbox{highlight}{highlighted}.

 \begin{figure*}[ht]
	\centering
	\includegraphics[width=0.95\textwidth]{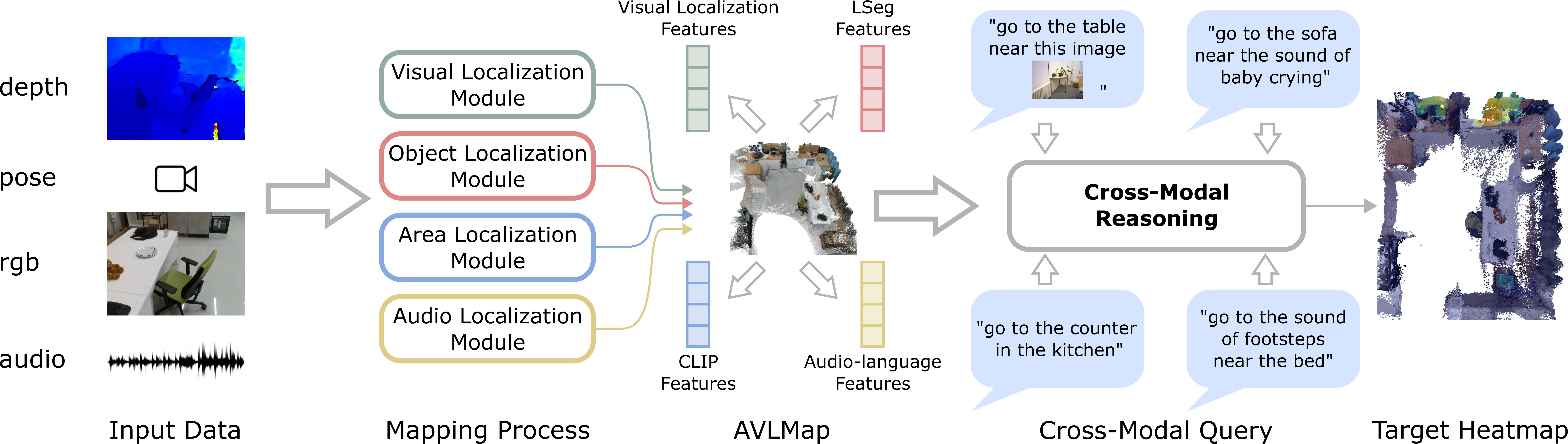}
	\caption{\small System overview. AVLMaps are constructed from RGB-D, audio, and odometry inputs, converting raw data into visual localization features, visual-language features, and audio-language features. During inference time, each module's output is unified with cross-modal reasoning, allowing users to query spatial location with multimodal information.}
	\label{fig:system_overview}

\end{figure*}

\subsection{Building an Audio Visual Language Map}
\label{subsec_building_avlmaps}
VLMaps provide an intuitive interface for humans to issue natural language commands to robots, enabling open-vocabulary spatial goal navigation. However, interacting with the world is inherently a multimodal experience. Since VLMaps primarily rely on visual information for map construction, it is important to consider other complementary sources of information that can enhance navigation. Motivated by this, we broaden the scope of VLMaps and propose a more general framework, AVLMaps, which enables robots to integrate and interpret multimodal information, including audio, images, and language, within a unified representation (as illustrated in Fig.~\ref{fig:system_overview}). In this paper, we present the extension to audio and image modalities as a case study, showcasing the modularity and extensibility of our multimodal spatial language maps. The AVLMaps framework is flexible and designed to accommodate future integration of additional sensing modalities, such as temperature, tactile feedback, or magnetic fields. The key idea behind AVLMaps is to combine visual localization features, pre-trained visual-language features, and audio-language features with a 3D reconstruction. Given an RGB-D video stream with an audio track and odometry information, we utilize four modules to build a multimodal features database. Given a specific query, each module returns predicted spatial locations on the map in the form of 3D voxel heatmaps. A heatmap can be denoted as $\mathcal{H} \in [0, 1]^{\bar{H} \times \bar{W} \times \bar{Z}}$, where $\bar{H}$, $\bar{W}$ and $\bar{Z}$ represent the size of the voxel map and the value in each element represents the probability of being the target position. $\mathbf{p} = (x, y, z)^T, \{x,y,z \in \mathbb{Z}\mid 1 \le x \le \bar{H}, 1 \le y \le \bar{W}, 1 \le z \le \bar{Z}\}$ is a voxel position in the map $\mathcal{H}$.

\begin{figure}[ht]
	\centering
	\includegraphics[width=1\columnwidth]{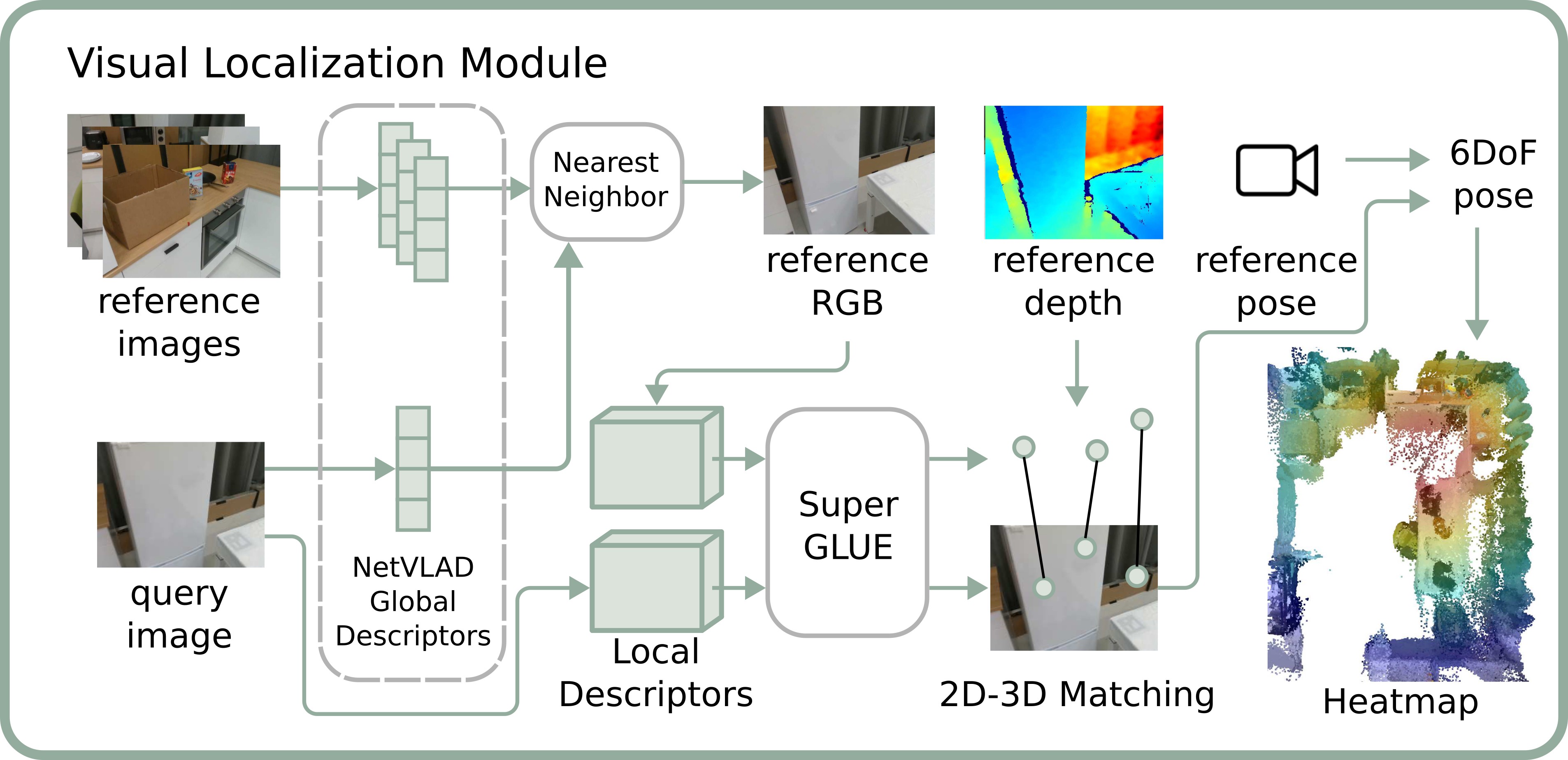}
	\caption{\small The overview of the visual localization module. We follow the hierarchical localization scheme by using NetVLAD and SuperGLUE to localize the query image's location before generating a heatmap as the indexing result.}
	\label{fig:visual_localization}

\end{figure}

\noindent\textbf{Visual Localization Module.} The overview of the module is shown in Fig.~\ref{fig:visual_localization}. The main purpose of this module is to localize a query image in our map. To achieve this goal, we follow a hierarchical localization scheme~\citep{sarlin2019coarse,sarlin2020superglue}. We first compute the NetVLAD~\citep{arandjelovic2016netvlad} global descriptors and SuperPoint~\citep{detone2018superpoint} local descriptors for all images from the RGB stream during exploration and store them with the corresponding depth and odometry as a reference database. During inference, we compute the global and local descriptors for the query image in the same manner. By searching the nearest neighbor of the query NetVLAD features in the reference database, we can find a reference image as our candidate. Next, we use SuperGLUE~\citep{sarlin2020superglue} to establish key point correspondences between the query image and the reference image we retrieve with NetVLAD from the database with their local SuperPoint features. With registered depth, we back-project the reference image's key points into the 3D space and obtain the 3D-2D correspondences for the query key points. In the end, we can apply the Perspective-n-Point method~\citep{fischler1981random} to estimate the query camera pose relative to the reference camera, and thus obtain the global camera pose with the odometry of the reference camera. 

In the visual localization module, the predicted global camera location is denoted as $\mathbf{p}_v = (x_v, y_v, z_v)^T$. In the heatmap $\mathcal{H}_v$, we define the probability at $\mathbf{p}_v$ as 1.0, and the probability linearly decays around this location according to the distance on the top-down map:
\begin{equation}
\label{eq:vis_loc_heatmap}
\mathcal{H}_{v}(\mathbf{p}) = \mathrm{max} (1.0 - \epsilon \cdot dist_{xy}(\mathbf{p}, \mathbf{p}_v), 0)
\end{equation}
\begin{equation}
\label{eq:dist_xy}
dist_{xy}(\mathbf{p}, \mathbf{q}) = \sqrt{(p_x - q_x)^2 + (p_y - q_y)^2}
\end{equation}
where $\epsilon$ is the decay rate, and $dist_{xy}(\mathbf{p}, \mathbf{q})$ denotes the distance between 3D vectors $\mathbf{p}$ and $\mathbf{q}$ on the $xy$-plane.

\begin{figure}[t]
	\centering
        \includegraphics[width=1\columnwidth]{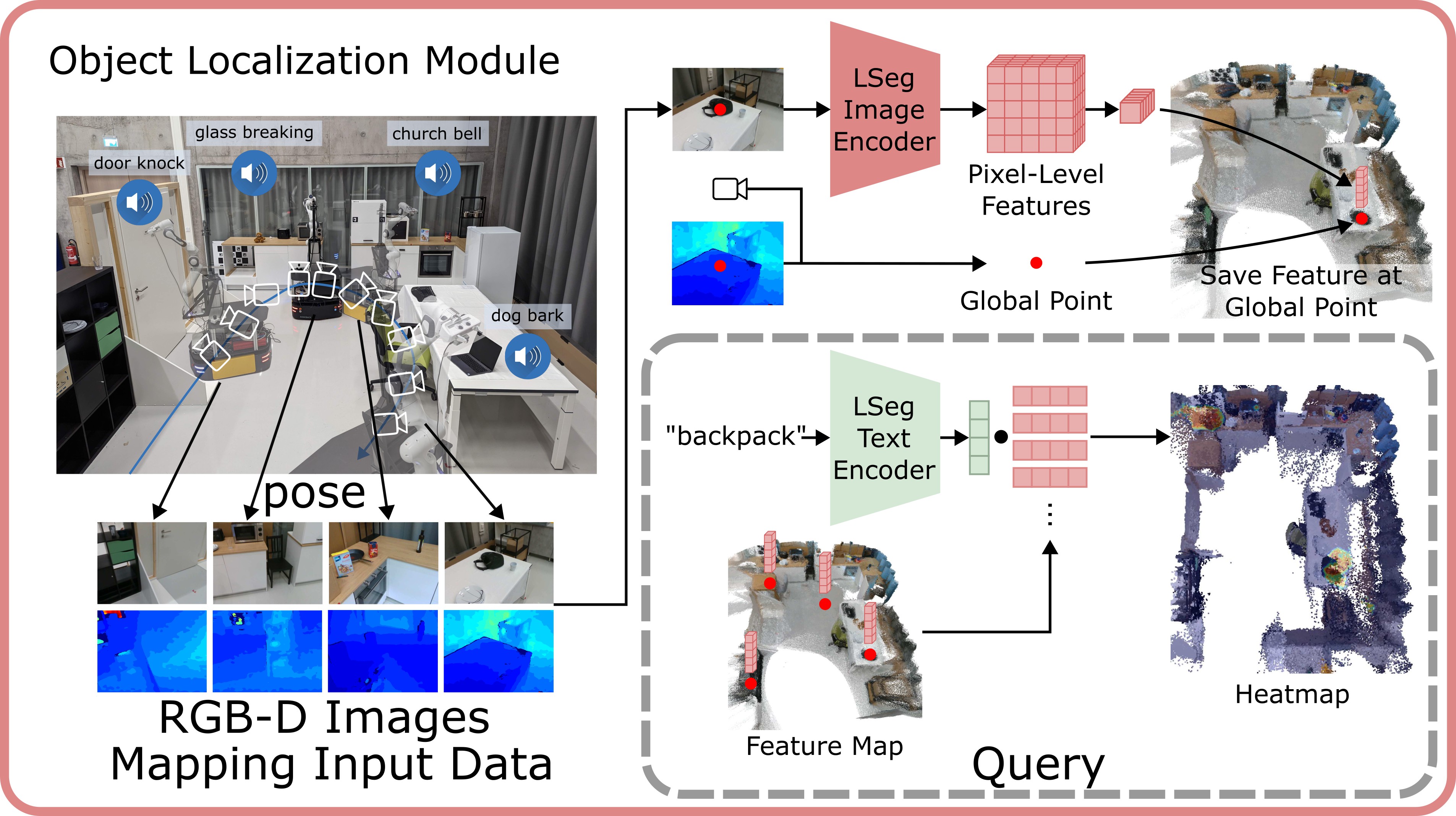}
	\caption{\small The overview of the object localization module. Similar to Sec.~\ref{sec:sub_map_creation}, during mapping, the RGB images in the exploration video are input to a vision-language model, LSeg~\citep{li2021language}, to generate pixel-level features. Corresponding pixels are back-projected with depth images and transformed to locations in the global coordinate frame, where the features are associated with. During inference, the query text is encoded by LSeg's text encoder, and a dot product with the point-level embeddings generates a score for each point. A heatmap is then created based on point scores and distances.}
	\label{fig:object_localization}

\end{figure}

\noindent\textbf{Object Localization Module.} The overview of the object localization module is shown in Fig.~\ref{fig:object_localization}. This module is abstracted from the VLMaps we introduced in Sec.~\ref{sec:sub_map_creation}, Sec.~\ref{sec:sub_landmark_indexing}, and Sec.~\ref{sec:obstacle_maps} with some minor changes. The key idea is to exploit an open-vocabulary segmentation method (\eg{} LSeg~\citep{li2022languagedriven} or OpenSeg~\citep{ghiasi2021open}) for pixel-level feature generation from the RGB image and to associate these features with the back-projected depth pixels in the 3D reconstruction. Different from Sec.~\ref{sec:sub_map_creation}, we don't project the 3D points into the top-down plane to create a 2D grid map but maintain a 3D voxel map where each voxel is associated with a visual-language feature. When there are multiple points projected into the same voxel, we store their mean features at the voxel. The inference process is similar to Sec.~\ref{sec:sub_landmark_indexing}. We define a list of categories in natural language and encode them with the language encoder. We compute the cosine similarity scores between all voxel-wise features and language features and use an $\mathrm{argmax}$ operator to select the top-scoring voxels for a certain category in the map. Depending on the application, the top-scoring 3D voxel points for a certain category can be used as the target point cloud for manipulation tasks or can be projected onto a top-down map for navigation purposes.

The object localization results are a list of points, denoted as $\{\mathbf{p}_{oi} = (x_{oi}, y_{oi}, z_{oi}) \mid  i = 1, \ldots, N\}$ where $N$ is the total number of points for the target object. We define the probabilities for all these locations as 1.0 in heatmap $\mathcal{H}_o$, and the probability linearly decays around these locations based on the Euclidean distance:
\begin{equation}
\label{eq:min_d}
d_{min} (\mathbf{p}) = \mathrm{min} \{dist(\mathbf{p}, \mathbf{p}_{oi}) \mid  i = 1, \ldots, N\}
\end{equation}
\begin{equation}
\label{eq:obj_loc_heatmap}
\mathcal{H}_{o}(\mathbf{p}) = \mathrm{max} (1.0 - \epsilon \cdot d_{min} (\mathbf{p}), 0)
\end{equation}
where $d_{min} (\mathbf{p})$ denotes the minimal distance between $\mathbf{p}$ and all object points $\{\mathbf{p}_{oi} \mid  i = 1, \ldots, N \}$, $dist (\mathbf{p}, \mathbf{q})$ denotes the Euclidean distance between $\mathbf{p}$ and $\mathbf{q}$.

In AVLMaps, the object localization module is also used to generate a 2D obstacle grid map for path planning, as is shown in Sec.~\ref{sec:obstacle_maps}. Different from the 2D feature grid map of a VLMap, the visual language map is now in the form of a 3D voxel grid where each occupied voxel is associated with a feature. We prompt the map with a list of free area concepts (\eg{}``floor'') and obstacle concepts (\eg{}``chair'', ``table'', ``counter'', and ``other'') for a specific embodiment, assigning a score to each concept for every voxel. Each voxel is then labeled with the concept that achieves the highest score, resulting in a 3D semantic voxel map. We merge the voxels labeled with obstacle concepts into a combined obstacle map, and then perform a top-down projection to produce a 2D obstacle grid map. In this grid, all pixel locations corresponding to projected obstacle voxels are marked as occupied, while all others are marked as free and navigable. It is worth noting that beyond object categories, AVLMaps also support the definition of audio or images as obstacles. For example, we can define ``the sound of glass breaking'', or a list of images as obstacles, generate their corresponding heatmaps, and treat locations with heat above a certain threshold as obstacles. This capability is useful when the forbidden regions are hard to describe with only object descriptions, like a glass-breaking scene without surrounding objects, or a region specified with a cell phone video.

\noindent\textbf{Area Localization Module.} While the object localization module is good at extracting object segments on the map, it falls short of localizing coarser goals such as regions (\eg{}``the area of the kitchen''). This is because the visual encoder for generating pixel-aligned features is obtained by fine-tuning a pre-trained model on a segmentation dataset, leading to the notorious catastrophic forgetting effect. Therefore, the visual encoder is better at segmenting common objects while worse at recognizing general visual concepts~\citep{jatavallabhula2023conceptfusion}. To take advantage of both pre-trained and fine-tuned methods, we propose to build an extra sparse topological CLIP features map similar to~\citep{shah2022lm}. The idea is to compute the CLIP visual features~\citep{radford2021learning} for all images from the RGB stream and associate the features with corresponding poses. During inference, given the language concept like ``the area of a bedroom'', we compute the language features with the CLIP language encoder and the image-to-language cosine similarity scores. These similarity scores indicate how likely these images match the language description. The odometry together with the score of each image indicates the predicted location with a confidence value.

 \begin{figure}[t]
	\centering
	\includegraphics[width=1\columnwidth]{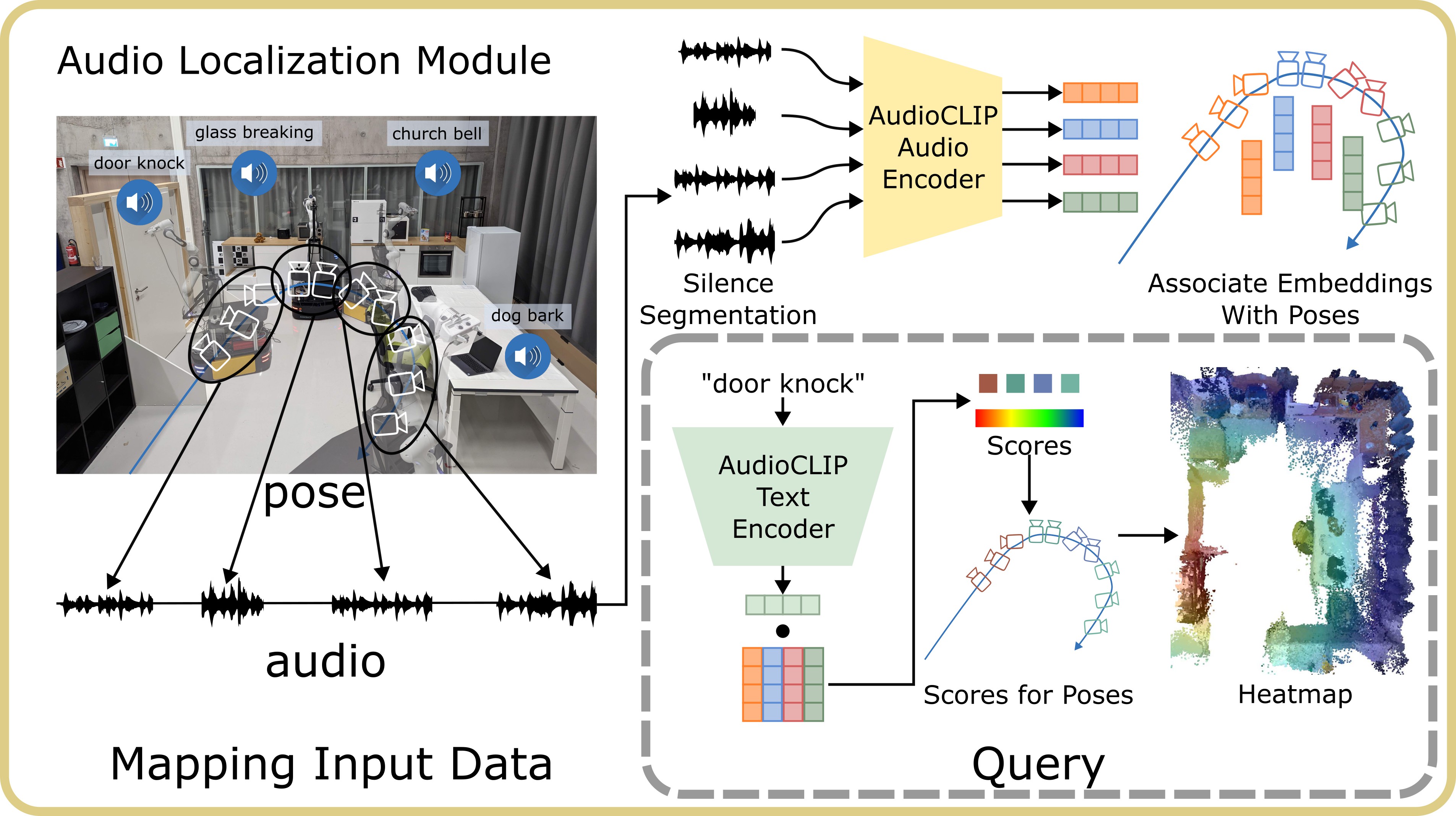}
	\caption{\small The overview of the audio localization module. During mapping, the exploration video's audio is segmented by silence and encoded using AudioCLIP's audio encoder. The resulting embeddings are linked to poses based on their timestamps. During inference, the query text is encoded by AudioCLIP's text encoder, and a dot product with the pose embeddings generates a score for each pose. A heatmap is then created based on pose scores and distances.}
	\label{fig:audio_localization}

\end{figure}

The area localization results are a list of position-confidence pairs, denoted as $\{(\mathbf{p}_{ai}, s_{ai})  \mid i = 1, \ldots, M\}$ where $M$ is the total number of frames in the input RGB-D stream. The scores $s_{ai}$ are normalized between 0 and 1. We define the probability for each point $\mathbf{p}_{ai}$ on the heatmap $\mathcal{H}_a$ as its score $s_{ai}$, and the probability linearly decays around the point on the $xy$-plane direction:

\begin{equation}
\label{eq:area_loc_heatmap_1}
\bar{\mathcal{H}}_{a}(\mathbf{p}) = \max_{i = 1, \ldots, M} \{s_{ai} - \epsilon \cdot dist_{xy}(\mathbf{p}, \mathbf{p}_{ai})\}
\end{equation}

\begin{equation}
\label{eq:area_loc_heatmap_2}
\mathcal{H}_{a}(\mathbf{p}) = \max (\bar{\mathcal{H}}_{a}(\mathbf{p}), 0)
\end{equation}
where the $\mathrm{max}$ operator for the curly brackets means taking the highest probability when a location is inside the affected regions for several $\mathbf{p}_{ai}$.

\noindent\textbf{Audio Localization Module.} The overview of the module is shown in Fig.~\ref{fig:audio_localization}. In this module, we utilize the audio information from the input stream. The key idea is to compute the audio-lingual features with audio-language pre-trained models such as wav2clip~\citep{wu2022wav2clip}, AudioCLIP~\citep{guzhov2022audioclip} or CLAP~\citep{elizalde2023clap}. We first segment the whole audio clip into several segments with silence detection. Whenever the volume is above a threshold, we mark this time step as the starting point of a segment. Whenever the volume of the sound is not larger than this threshold for a certain duration, we end the segment. In the next step, we compute the audio features for each segment with pre-trained audio-language models and associate the features with the odometry at the specific segment. During inference, given a language description of the sound, like ``the sound of door knocks'', we encode the language into language features and compute the matching scores between the language and all audio segments in the same way as in the object localization module. The odometry associated with the top-scoring segment is the predicted location.

The audio localization results are similar to those of the area localization module. The position-score pairs are denoted as $\{(\mathbf{p}_{si}, s_{si}) \mid i = 1, \ldots, K\}$ where $K$ is the total number of sound segments in the input video stream. The heatmap $\mathcal{H}_s$ is defined as:
\begin{equation}
\label{eq:sound_loc_heatmap_1}
\bar{\mathcal{H}}_{s}(\mathbf{p}) = \max_{i = 1, \ldots, K} \{s_{si} - \epsilon \cdot dist_{xy}(\mathbf{p}, \mathbf{p}_{si}) \}
\end{equation}

\begin{equation}
\label{eq:sound_loc_heatmap_2}
\mathcal{H}_{s}(\mathbf{p}) = \max (\bar{\mathcal{H}}_{s}(\mathbf{p}) - \epsilon \cdot dist_{xy}(\mathbf{p}, \mathbf{p}_{si}), 0)
\end{equation}

\subsection{Cross-Modality Reasoning}
\label{subsec_cross_modal_reasoning}
A key advantage of our method is its capability to disambiguate goals with additional information, even from different modalities. The goal of the cross-modality reasoning method is to output a target location of a specific concept (for example, ``the sofa'') given the information of other nearby concepts (for example, ``near the sound of glass breaking''). In the last section, we introduced how we generate heatmaps for concepts of different modalities. Given these heatmaps, we want to further narrow down the target location.

\begin{figure}[t]
	\centering
	\includegraphics[width=1\columnwidth]{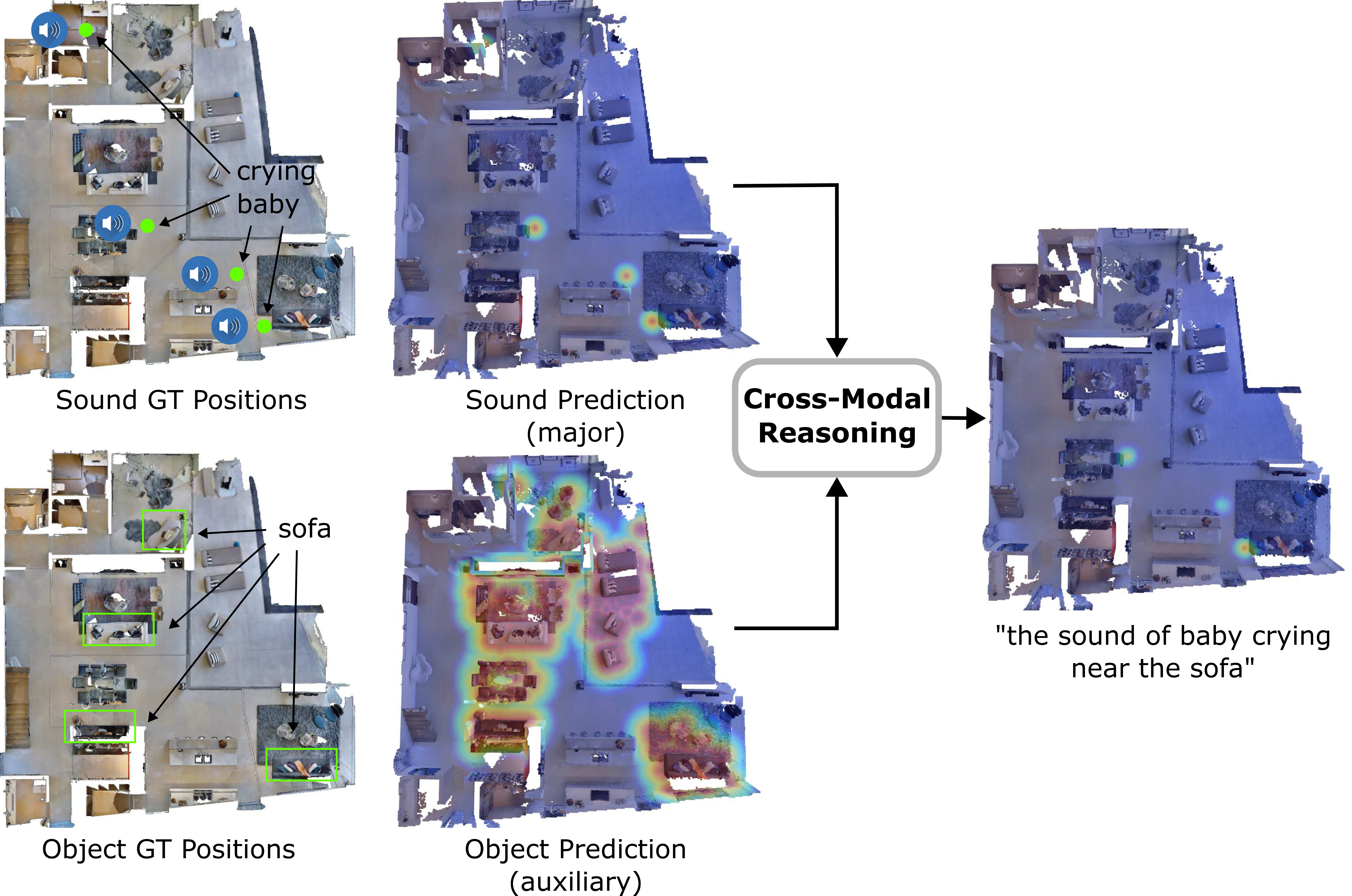}
	\caption{\small The key idea of cross-modal reasoning is converting the prediction from different modalities into heatmaps, and then fusing them with element-wise multiplication, effectively using complementary multimodal information to resolve ambiguous prompts.}
	\label{fig:cross_modal_fusion}

\end{figure}

\noindent\textbf{Cross-Modal Reasoning.} The main idea of our cross-modal reasoning method is shown in Fig.~\ref{fig:cross_modal_fusion}. We treat the predictions from four modules as four modalities. When there are several queries referring to different modalities, we compute the respective heatmaps first and then perform element-wise multiplication among all heatmaps:
\begin{equation}
\label{eq:heatmap_mul}
\mathcal{H}_{\mathit target} = \mathcal{H}_{1} \odot \ldots \odot \mathcal{H}_{L} 
\end{equation}
where $\odot$ is the element-wise multiplication operator, and $L$ is the total number of referred modalities. We extract the position on the target heatmap $\mathcal{H}_{target}$ that has the highest probability as the predicted location.

When we compute the heatmaps, there is always a primary heatmap while others are auxiliary ones. For example, in Fig.~\ref{fig:cross_modal_fusion}, in the query ``the sound of baby crying near the sofa'', the heatmap for ``the sound of baby crying'' is the primary heatmap, while the heatmap for ``the sofa'' is the auxiliary. We set the decay rate for the primary heatmap higher (\eg{}0.1 in this work for voxel map with 0.05 meter voxel size) since we want to know the exact location of the target while tuning the decay rate for the auxiliary heatmap lower (\eg{}0.01) as having a broader effect area to narrow down major targets is desirable. More specifically, a higher decay rate indicates that the relevancy to the concept decreases faster when the location is farther away from the concept locations  (heat value decreases 0.1 for every 0.05 meter). When there are multiple concepts or modalities mentioned in the target specification, the map with a high decay rate refers to the target concept we want the robot to get closer to. Those maps with low decay rates serve as constraints to select targets in the main map. From another perspective, the decay rates of multiple maps represent the importance weights we assign to their corresponding concepts. The higher the decay rate is, the more important that concept is, and we want the robot to get closer to that concept. When the decay rates of two concepts are the same, the two concepts are equally important, and the fusion of the two concepts' heatmaps will peak at the middle points between those concepts.

It is worth noting that our cross-modal reasoning method is relatively simple and has the underlying assumption that the heatmaps from different modalities are conditionally independent. Despite this simplification, our method proves to be effective across our experiments. Nonetheless, exploring more advanced cross-modal fusion techniques remains an exciting direction for future work.


\begin{table*}[!htp]
  \centering
  \begin{tabularx}{\textwidth}{l|X}
  \toprule

    APIs          & Functions                       \\
\midrule
\texttt{\textbf{load\_image}(path)}         & load an image as an array from the path.    \\
\rowcolor{highlight}
\texttt{\textbf{move\_to}(pos)}         & move to a position on the map.    \\
\texttt{\textbf{get\_major\_map}(img=None, obj=None, sound=None)}       &
generate a heatmap with high heat decay rate for a specific modality (image, object description, or sound description). \\
\rowcolor{highlight}
\texttt{\textbf{get\_map}(img=None, obj=None, sound=None)}       &
generate a heatmap with low heat decay rate for a specific modality (image, object description, or sound description). \\
\texttt{\textbf{get\_max\_pose\_3d}(heatmap)}       &
retrieve the coordinate in a 3d voxel heatmap which has the maximal heat value). \\
  \bottomrule
  \end{tabularx}
  \caption{Navigation API library for multimodal goal navigation in AVLMaps.}
  \label{table:nav_primitives}
\end{table*}

\subsection{Multimodal Goal Navigation from Language}\label{subsec_multimodal_nav_from_language}
In the setting of multimodal goal navigation from language, the agent is given language descriptions referring to targets from different modalities (\eg{}sounds, images and objects), and is required to plan paths to them. While most of the previous navigation methods focus mainly on a specific type of goal, we unify these tasks with the help of large language models (LLMs). Following a similar spirit to Sec.~\ref{sec:sub_language_spatial_navigation}, we use an LLM to interpret the natural language commands and synthesize API calls combined with simple logic structures in the form of executable python code. In the following, we will detail the code generation process for multimodal goal navigation, including (i) the introduction of the API library as a tool set for the LLM to use during code generation, (ii) the mechanism of spatial goal reasoning, (iii) the way we generate multimodal maps with code, (iv) the method to achieve cross-modal reasoning with code, and (v) the conversion of code to navigation commands that robots receive. At the end, we provide the prompt and an example of inference for our method.

\noindent\textbf{Navigation API library. }
In Table~\ref{table:nav_primitives}, we listed all the APIs we provide to the LLM for potential usage. Compared to VLMaps' API library in Table~\ref{table:spatial_nav_primitives}, we simplify and adapt the library to more modalities for AVLMaps. We provide a list of examples consisting of language instructions and generated code, which demonstrates the usage of those APIs as a context prompt (as in Fig.~\ref{fig:full_prompt}) to the LLM when we ask it to generate code during the inference time.

\begin{figure}[htp]
    \centering
    \lmp{
    \prompt{
    \# move to the middle of the sound of cat meowing and\\
    \# the image: /path/to/image.png\\
    img = robot.load\_image("/path/to/image.png")\\
    sound\_map = robot.get\_major\_map(sound="cat meowing")\\
    img\_map = robot.get\_major\_map(img=img)\\
    pos1 = robot.get\_max\_pos\_3d(sound\_map)\\
    pos2 = robot.get\_max\_pos\_3d(img\_map)\\
    pos = (pos1 + pos2) / 2\\
    robot.move\_to(pos)\\
    \\
    \# move to the window next to the sound of\\
    \# glass breaking\\
    obj\_map = robot.get\_major\_map(obj="window")\\
    sound\_map = robot.get\_map(sound="glass breaking")\\
    fuse\_map = obj\_map * sound\_map\\
    pos = robot.get\_max\_pos\_3d(fuse\_map)\\
    robot.move\_to(pos)\\
    \\
    \# move to the sound of crying baby next to the \\
    \# counter\\
    obj\_map = robot.get\_map(obj="counter")\\
    sound\_map = robot.get\_major\_map(sound="crying baby")\\
    fuse\_map = obj\_map * sound\_map\\
    pos = robot.get\_max\_pos\_3d(fuse\_map)\\
    robot.move\_to(pos)\\
    \\
    \# move to the table next to the sound of \\
    \# crying baby and the sound of dog\\
    obj\_map = robot.get\_major\_map(obj="table")\\
    sound\_map = robot.get\_map(sound="crying baby") / \\
    \ \ \ \ \ \ * robot.get\_map(sound="dog")\\
    fuse\_map = obj\_map * sound\_map\\
    pos = robot.get\_max\_pos\_3d(fuse\_map)\\
    robot.move\_to(pos)\\
    \\
    \# move to the middle of the table and \\
    \# the chair next to the sound of crying baby\\
    obj\_map\_1 = robot.get\_major\_map(obj="table")\\
    obj\_map\_2 = robot.get\_major\_map(obj="chair")\\
    sound\_map = robot.get\_map(sound="crying baby")\\
    fuse\_map = obj\_map\_2 * sound\_map\\
    pos1 = robot.get\_max\_pos\_3d(obj\_map\_1)\\
    pos2 = robot.get\_max\_pos\_3d(fuse\_map)\\
    pos = (pos1 + pos2) / 2\\
    robot.move\_to(pos)
    }}
    \caption{The full context prompt (prompt in {\color{prompt-gray}gray}) AVLMaps used for achieving all navigation tasks in the experiments.}
    \label{fig:full_prompt}
\end{figure}

\begin{figure}[htp]
    \centering
    \speciallmp{
    \command{\# move in between the image ./006899.png and the}\\
    \command{backpack near the sound of glass breaking}\\
    \hlcode{img = robot.load\_image("./006899.png")}\\
    \hlcode{img\_map = robot.get\_major\_map(img=img)}\\
    \hlcode{obj\_map = robot.get\_major\_map(obj="backpack")}\\
    \hlcode{sound\_map = robot.get\_map(sound="glass breaking")}\\
    \hlcode{fuse\_map = obj\_map * sound\_map}\\
    \hlcode{pos1 = robot.get\_max\_pos\_3d(img\_map)}\\
    \hlcode{pos2 = robot.get\_max\_pos\_3d(fuse\_map)}\\
    \hlcode{pos = (pos1 + pos2) / 2}\\
    \hlcode{robot.move\_to(pos)}
    }
    \caption{The query and the generated results from the LLM. During the query, the context prompt in Fig.~\ref{fig:full_prompt} and the input task commands are prompted to the LLM together. The input task commands are in \command{green} and generated outputs are \colorbox{highlight}{highlighted}}
    \label{fig:query_prompt}
\end{figure}

\noindent\textbf{Spatial Goal Reasoning. }
We follow the intuition in Sec.~\ref{sec:sub_language_spatial_navigation} that spatial locations can be computed with simple math during code generation. For example, the location of ``in between the counter and the fridge'' can be obtained by getting the positions of the counter and the fridge respectively and apply an average to the two locations. In the multimodal goal navigation prompt, we reduce the spatial goal API calling examples to lay the focus more on multimodal targets and cross-modal reasoning. In principle, more diverse spatial concept reasoning examples could be integrated into the prompt, as is shown in Sec.~\ref{sec:sub_language_spatial_navigation}.

\noindent\textbf{Multimodal Heatmap Generation with Code. }
For heatmap generation, we implement interfaces \texttt{\textbf{get\_major\_map}(obj=None, sound=None, img=None)} and \texttt{\textbf{get\_map}(obj=None, sound=None, img=None)} (Table~\ref{table:nav_primitives}). They take an object name, a sound name, or an image as input and output heatmaps indicating the locations of targets. These two functions are implemented following the localization modules in Sec.~\ref{subsec_building_avlmaps}. The \texttt{\textbf{get\_major\_map}} generates heatmaps with a higher decay rate while \texttt{\textbf{get\_map}} creates heatmaps with a lower decay rate. To support the image prompt, we add the image path in the language query like ``the image /path/to/image.png'' and use LLMs to call the image loading APIs.

\noindent\textbf{Cross-Modal Reasoning with Code. }
As is introduced before, the logic of the cross-modal reasoning is relatively simple, which is just an element-wise multiplication of all relevant heatmaps. In the code, it can be performed with one line of code \colorbox{highlight}{\texttt{fusemap = map1 $*$ map2}}.

\noindent\textbf{Navigation Commands from Code. }
The API \texttt{move\_to(pos)} takes a 3D voxel grid position as input, projects it onto the 2D grid map. It applies a planning algorithm on the grid map to generate a path on the map. The points on the path are treated as a list of subgoals and are used to generate a list of low-level actions (the action space contains turning left or right 5 degrees and moving forward 0.25 meter) to reach them sequentially. In the end, the list of low-level action commands is executed to finish the navigation instruction.

\noindent\textbf{Prompt and an Inference Example.} Since large language models are great few-shot learners~\citep{brown2020language}, they are able to learn and imitate internal patterns when several simple examples are provided as context in addition to the direct query. To enable the LLM to understand how to use our APIs, we need to provide a few examples of how to use these APIs to tackle tasks described with language instructions. During the inference, we prompt the full context prompts (in {\color{prompt-gray}gray}) in Fig.~\ref{fig:full_prompt} together with task command (in \command{green}) in Fig.~\ref{fig:query_prompt} and an example of the generated outputs are \colorbox{highlight}{highlighted}. In our work, we use OpenAI's \texttt{text-davinci-003} model as our LLM for all experiments.

\section{Experiments}
\label{sec:experiments}

In this section, we aim to evaluate our multimodal spatial map representations in a variety of tasks. More specifically, we address nine key questions: (i) how is VLMaps' spatial language goals navigation performance compared to recent open-vocabulary navigation baselines (Sec.~\ref{sec:exp_spatial_goal_navigation_language}), (ii) whether VLMaps with their capacity to specify open-vocabulary obstacle
maps can provide utility in improving the navigation efficiency of
different robot embodiments (Sec.~\ref{sec:exp_multi_embodiment_navigation}), (iii) how AVLMaps enable a robot to navigate to multimodal goals, including sound, image, and object queries (Sec.~\ref{sec:exp_multi_modal_goal_nav}), (iv) how our cross-modality reasoning approach helps a robot to disambiguate goals with multimodal information (Sec.~\ref{sec:exp_cross_modal_goal_indexing}), (v) how the performance of AVLMaps scales with recent advanced foundation models specialized in different modalities (Sec.~\ref{sec:exp_scale}), and (vi) how AVLMaps' multimodal indexing and reasoning capabilities translate to real-world environments, empowering robots with diverse embodiments to perform tasks that demand comprehensive multimodal understanding, such as mobile navigation (Sec.~\ref{sec:exp_real_world}) and table-top manipulation with multimodal prompts (Sec.~\ref{sec:exp_table_top}).

\subsection{Zero-Shot Spatial Goal Navigation from Language}
\label{sec:exp_spatial_goal_navigation_language}

\noindent\textbf{Experimental setup.} We use the Habitat simulator~\citep{habitat19iccv} with the Matterport3D dataset~\citep{Matterport3D} for the evaluation of multi-object and spatial goal navigation tasks. The dataset contains a large set of realistic indoor scenes that help evaluate the generalization capabilities of navigating agents. To evaluate the creation of open-vocabulary multi-embodiment obstacle maps, we adopt the AI2THOR simulator due to its support of multiple agent types, such as LoCoBot and drone. In these two environments, the robot is required to navigate in a continuous environment with actions: \textbf{move forward 0.05 meters}, \textbf{turn left 1 degree}, \textbf{turn right 1 degree} and \textbf{stop}. For map creation in Habitat, we collect 12,096 RGB-D frames across ten different scenes and record the camera pose of each frame. 
\\
\\
\noindent\textbf{Baselines.} We evaluate VLMaps against three baseline methods, all of which utilize visual-language models and are capable of zero-shot language-based navigation:
\begin{itemize}[leftmargin=*]
  \item LM-Nav~\citep{shah2022lm} creates a graph where image observations of an environment are stored as nodes while the proximity between images are represented as edges. By combining GPT-3 and CLIP, it parses language instructions into a list of landmarks and plans on the graph towards corresponding nodes. 
  \item CLIP on Wheels (CoW)~\citep{gadre2022clip} achieves language-based object navigation by building a saliency map for the target category with CLIP and GradCAM~\citep{selvaraju2020grad}. By thresholding the saliency values, it retrieves a segmentation mask for the target object category and then plans the path on the map.
  \item CLIP-features-based map (CLIP Map) is an ablative baseline that generates a feature map for the environment in a similar way as ours. Instead of using LSeg visual features, it projects the CLIP visual features onto the map averaged across views. Object category masks are generated by thresholding the similarity between map features and the object category features.
\end{itemize}
For additional context and analysis, we also report results from a system that has access to a ground truth semantic map for navigation (GT Map), to provide a systems-level upper bound on performance.
\\
\\
\noindent\textbf{Tasks Collection.} In these experiments, we investigate the performance of VLMaps versus other baselines for zero-shot \textit{spatial} goal navigation from language. Our benchmark consists of 21 trajectories in seven scenes, with manually specified corresponding language instructions for evaluation. Each trajectory contains four different spatial locations as subgoals. Examples of subgoals are ``east of the table'', ``in between the chair and the sofa'', or ``move forward 3 meters''. There are also instructions for the robot to realign itself in reference to nearby objects such as ``with the counter on your right''. We only consider a subgoal as having been achieved, when the robot reaches the subgoal location within a range of one meter. To evaluate the long-horizon navigation capabilities of the agents, we compute the success rate (SR) of continuously reaching one to four subgoals in a sequence, shown in Tab.~\ref{table:spatial_lang_nav}. For all map-based methods, including CoW, CLIP Map, ground truth semantic map, and our method, we apply the code generation techniques introduced in Sec.~\ref{sec:sub_language_spatial_navigation}. For LM-Nav, we simply use the same parsing method in the original paper~\citep{shah2022lm} to break down the language instruction into subgoals.

\begin{table}[h]
  \centering

  \begin{tabularx}{\linewidth}{l>{\centering\arraybackslash}X>{\centering\arraybackslash}X>{\centering\arraybackslash}X>{\centering\arraybackslash}X>{\centering\arraybackslash}X>{\centering\arraybackslash}X>{\centering\arraybackslash}X>{\centering\arraybackslash}X}
  
  \toprule
\multirow{2}[1]{*}{Tasks}  & \multicolumn{4}{c}{No. Subgoals in a Row}       \\                
                    \cmidrule(lr){2-5}
                   & 1             & 2           & 3           & 4             \\
\midrule
LM-Nav \citep{shah2022lm}& 5                  & 5               & 0               & 0                        \\
CoW \citep{gadre2022clip}& 33                  & 5               & 0               & 0                        \\
CLIP Map           & 19                  & 0                & 0               & 0                        \\
VLMaps (ours)    & \textbf{62}         & \textbf{33}      & \textbf{14}     & \textbf{10}          \\
\midrule
GT Map    & 76        & 48      & 33     & 29          \\

  \bottomrule
  
  \end{tabularx}

    \caption{The success rate (\%) of zero-shot spatial goal navigation with language.}

\label{table:spatial_lang_nav}
\end{table}

Tab.~\ref{table:spatial_lang_nav} summarizes the zero-shot spatial goal navigation success rates. Our method outperforms other baselines in this task. Different from object navigation tasks where agents only need to approach a certain object type within a range, disregarding the relative spatial shift to the object, the language-based spatial goal navigation tasks require the robot to accurately arrive at the described location in reference to the object. This poses a bigger challenge to the landmark localization ability of the method. The low localization ability of CoW and CLIP Map leads to their high failure rates in this task.

\begin{figure*}[!th]
	\centering
	\includegraphics[width=1\textwidth]{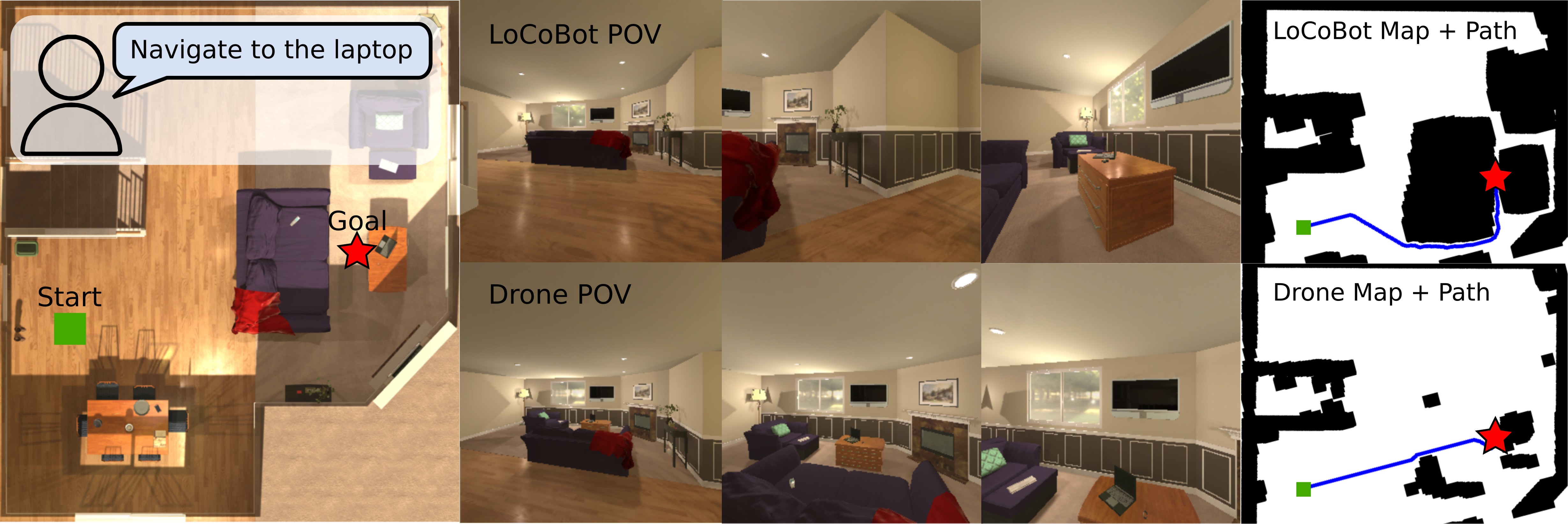}
	\caption{VLMaps enable different embodiments to define their own obstacle maps for navigation. The left image shows the top-down view of an environment. The middle columns show the observations of agents during navigation. The images on the right demonstrate the obstacles maps generated for different embodiments and the corresponding navigation paths.}
	\label{fig:multi_embodiment}

\end{figure*}

\begin{table*}[h]
  \setlength\tabcolsep{1pt}
  \centering
  \footnotesize

  \begin{tabularx}{\textwidth}{l>{\centering\arraybackslash}X>{\centering\arraybackslash}X>{\centering\arraybackslash}X>{\centering\arraybackslash}X>{\centering\arraybackslash}X>{\centering\arraybackslash}X>{\centering\arraybackslash}X>{\centering\arraybackslash}X>{\centering\arraybackslash}X>{\centering\arraybackslash}X}
  \toprule
\multirow{3}[3]{*}{Tasks}  & \multicolumn{8}{c}{No. Subgoals in a Row} &  Indep.      \\                
                    \cmidrule(lr){2-9}
        & \multicolumn{2}{c}{1}  & \multicolumn{2}{c}{2} & \multicolumn{2}{c}{3} & \multicolumn{2}{c}{4} &  Subgoals\\
         \cmidrule(lr){2-3}  \cmidrule(lr){4-5}  \cmidrule(lr){6-7}  \cmidrule(lr){8-9}  \cmidrule(lr){10-10}
        & SR      & SPL      & SR      & SPL      & SR      & SPL      & SR      & SPL      & SR      \\
\midrule
LoCoBot (ground map)     & 53      & \textbf{49.0}      & 28      & \textbf{17.8}       & 14      & 6.7       & 6       & \textbf{2.5}        & 52.3      \\
Drone (ground map)       & 53      & 41.8       & 28      & 15.5       & 14      & 5.3       & 6       & 2.0        & 53.3      \\
Drone (drone map)        & \textbf{56} & 45.4 & \textbf{30} & 16.3 & \textbf{17} & \textbf{7.0} & \textbf{7} & \textbf{2.5} & \textbf{55.0}      \\

  \bottomrule
  \end{tabularx}

  \caption{The success rate (\%) and SPL of multi-embodiment object goal navigation with language.}

  \label{table:multi_embodiment_nav}
  
\end{table*}

\subsection{Cross-Embodiment Navigation}
\label{sec:exp_multi_embodiment_navigation}

\noindent\textbf{Experimental Setup.} We collect 1,826 RGB-D frames across ten rooms in AI2THOR~\citep{kolve2017ai2} and build the VLMaps for these scenes. We study the ability of VLMaps to improve navigation efficiency by retrieving different obstacle maps for different embodiments (given the same VLMap) in navigation tasks. We evaluate more than 100 sequences of object subgoals in the AI2THOR simulator. We evaluate VLMaps on both a LoCoBot and a drone to test its capability of generating obstacle maps at runtime for multi-embodiment navigation.

\noindent\textbf{Obstacle Maps Generation.} We apply the open-vocabulary obstacle map generation method in Sec.~\ref{sec:obstacle_maps} to create an obstacle map for the drone (drone map) and one for the LoCoBot (ground map) by defining obstacles for them differently. For the LoCoBot (ground robot), we first define a potential obstacle list as \texttt{[``chair'', ``wall'', ``wall above the door'', ``table'', ``window'', ``floor'', ``stairs'', ``other'']} and perform open-vocabulary landmark indexing. Later, we only select the union of the masks for the objects \texttt{``wall'', ``chair'', ``table'', ``window'', ``stairs'', ``other''} as the obstacle map. For the drone (flying robot), we perform landmark indexing with the potential obstacle list: \texttt{[``chair'', ``sofa'', ``wall'', ``table'', ``counter'', ``window'', ``floor'', ``stairs'', ``ceiling lights'', ``cabinet'', ``counter support'', ``other'']}. Afterwards, we take union of the masks for \texttt{[``wall'', ``window'', ``stairs'', ``ceiling lights'', ``cabinet'', ``other'']} to generate the obstacle map.
\\
\\
\noindent\textbf{Baselines.} We test the navigation ability of these embodiments with three setups: a LoCoBot with a ground map, a drone with a ground map, and a drone with a drone map.
\\
\\
\noindent\textbf{Metrics.} We evaluate the Success Rate (SR) and the Success rate weighted by the (normalized inverse) Path Length (SPL) \citep{anderson2018evaluation} defined as: $SPL = \frac{1}{N} \sum_{i=1} ^{N} S_i \frac{\textit{l}_i}{max(\textit{p}_i, \textit{l}_i)}$
where $N$ is the total number of evaluated tasks, $S_i \in \{0, 1\}$ is the binary indicator of success, $\textit{l}_i$ denotes the ground truth shortest path length, and $\textit{p}_i$ denotes the actual path length of the agent in navigation. This metric indicates how efficient the actual path is compared to the ground truth shortest path when the navigation task is achieved. In our three setups, the ground truth trajectories for the LoCoBot and the drone are planned on floor-level and on height level of 1.7 meters respectively.
\\
\\
The results provided in Tab.~\ref{table:multi_embodiment_nav} show that the average navigation success rates of the ground-map version of the LoCoBot and the drone are similar because the same obstacles map is used for planning. However, there is an obvious gap between their SPL values. This is because when the drone does not have access to a customized obstacle map, it fails to benefit from flying over ground objects to improve the navigation efficiency. In contrast, while achieving similar success rate compared to the drone with a ground map, the drone with a drone map manages to navigate with higher path efficiency, reflected by the increased SPL values. The comparable SPL values for the drone with the drone map and the LoCoBot with the ground map shows that VLMaps help to generalize the navigation efficiency among different embodiments. An example of the multi-embodiment object navigation task is shown in Fig.~\ref{fig:multi_embodiment}, where by defining a more efficient obstacles map, the drone flies over the sofa and reaches the laptop target directly, while the LoCoBot has to move aside first to avoid colliding with the sofa.

\subsection{Multimodal Navigation Simulation Setup}\label{sec:exp_sim_setup}

\textbf{Exerimental setup.} We use the Habitat simulator~\citep{habitat19iccv,szot2021habitat} with the Matterport3D dataset~\citep{Matterport3D} for the evaluation of multimodal navigation tasks. For mapping purposes, we manually collect RGB-D video streams in the simulator across ten different scenes and add random audio tracks to the videos to simulate the audio sensing modality. All audio comes from the validation fold (Fold-1) of the ESC-50 dataset~\citep{piczak2015esc}, which contains 50 categories of common sounds. In navigation tasks, the robot has four actions to take: \textbf{move forward 0.1 meters}, \textbf{turn left 5 degrees}, \textbf{turn right 5 degrees}, and \textbf{stop}. In sequential goal setting, the robot is required to navigate to a sequence of goals and take the \textbf{stop} action when it reaches each subgoal. When the stop position is less than one meter away from the ground truth position, the subgoal is considered successfully finished.
\\
\\
\noindent\textbf{Tasks collection.} In multimodal goal navigation tasks in Sec.~\ref{sec:exp_multi_modal_goal_nav}, we consider three kinds of goals: image goals, object goals, and sound goals. For image goals, we randomly sample positions and orientations on the top-down map and render images as targets. For object goals, we access the metadata (\eg bounding boxes and semantics) from the Matterport3D dataset and sample a list of categories in each scene as queries. For sound goals, we randomly sample sound classes of audio merged with the mapping videos as targets, treating the video frame positions as the ground truth. 

In cross-modal goal indexing tasks in Sec.~\ref{sec:exp_cross_modal_goal_indexing}, we collect three types of datasets: 
\begin{itemize}[leftmargin=*]
  \item \textbf{Visual-Object cross-modal indexing} We manually select image-object pairs on the top-down map for localizing ``an object X near the image Y''.
  \item \textbf{Area-Object cross-modal indexing} We access the region and object metadata (\eg bounding boxes and semantics) from the Matterport3D dataset to automatically generate a list of object-region pairs. This dataset is for localizing ``an object X in the area of Y''.
  \item \textbf{Object-Sound cross-modal indexing} We manually insert several sounds of the same kind into a scene and select for each sound location a nearby object for disambiguation. The query is ``a sound X near the object Y''.
\end{itemize}

In cross-modal goal navigation in Sec.~\ref{sec:exp_cross_modal_nav}, we randomly sample starting pose in ten scenes and treat the visual-object and object-sound cross-modal goals in Sec.~\ref{sec:exp_cross_modal_goal_indexing} as navigation goals. For all cross-modal navigation and indexing tasks, we use the prompt introduced in Sec.~\ref{subsec_multimodal_nav_from_language} to generate navigation commands.

\subsection{Multimodal Goal Navigation}\label{sec:exp_multi_modal_goal_nav}
\textbf{Sound goal navigation.} We first test AVLMaps in sound goal navigation tasks. We collect 200 sequences of sound goals in ten different scenes. In each sequence, there are four sound categories that require the robot to reach. The results are shown in Tab.~\ref{table:multi_audio_nav}. We generate AudioCLIP~\citep{guzhov2022audioclip} features with our audio localization module and match all audio with the target sound category in the embedding space, similar to a text-to-audio retrieval setup. Then, the agent plans a path to the audio position. We tested different ranges of sound categories inserted into the map. The full list of sound categories in each major class can be found in the link\endnote{\href{https://github.com/karolpiczak/ESC-50}{https://github.com/karolpiczak/ESC-50}}. The results show that our agent manages to recognize sound goals and navigate with a 77.5\% success rate.
\begin{table}[h]
  \setlength\tabcolsep{2.5pt}
  \footnotesize\sf\centering
  \begin{tabular}{lccccc}
  \toprule
\multirow{2}[1]{*}{Tasks}  & \multicolumn{4}{l}{No. Subgoals in a Row} & Independent      \\                
                    \cmidrule(lr){2-5}
        &   1             & 2           & 3           & 4    &  Subgoals\\
\midrule
Domestic Sound (10 categories) & 59.5         & 33.0      & 15.5     & 7.0     & 62.5     \\
+ Human Sound (20 categories) & 69.5         & 47.0      & 36.5     & 23.0     & 72.38     \\
+ Animal Sound (30 categories) & 74.5         & 58.5      & 45.5     & 33.0     & 77.5     \\

  \bottomrule
  \end{tabular}
  \caption{The success rate (\%) of sound goal navigation with AVLMaps.}
  \label{table:multi_audio_nav}

\end{table}

\noindent\textbf{Visual and object goals navigation.} We then test AVLMaps with visual and object goal navigation tasks. The agent is given an image and two object categories in the language in one sequence of tasks and asked to navigate to the image goal and two object goals in sequence. The success rate for 200 sequences of tasks in ten scenes is reported in Tab.~\ref{table:visual_object_goal_nav}. The results show that our method enables the agent to navigate to goals from different modalities.
\begin{table}[h]
  \setlength\tabcolsep{7.4pt}
  \footnotesize\sf\centering
  \begin{tabular}{lccccc}
  \toprule
\multirow{2}[1]{*}{Tasks}  & \multicolumn{3}{l}{No. Subgoals in a Row} & Independent      \\                
                    \cmidrule(lr){2-4}
               &   1             & 2           & 3              &  Subgoals\\
\midrule
AVLMaps (Ours) & 71.5            & 40.5        & 25.0           & 47.4     \\

  \bottomrule
  \end{tabular}
  \caption{The success rate (\%) of multimodal goal navigation with AVLMaps. The agent is required to navigate to one visual goal, and two object goals in sequence.}
  \label{table:visual_object_goal_nav}

\end{table}

\subsection{Cross-Modal Goal Indexing}\label{sec:exp_cross_modal_goal_indexing}
When we refer to a goal with language, it is likely that the goal can be found in more than one place in the environment. A major strength of our method is that it can disambiguate goals with multimodal information. In this experiment, we will show the cross-modal goal reasoning capability of AVLMaps.
\\
\\
\noindent\textbf{Area-Object goal indexing.} In this setup, we use an area description to disambiguate the object goal. We collected 100 indexing tasks in ten scenes. Each task consists of an object category and a region category (\eg ``living room'', ``kitchen'', ``dining room'', ``bathroom'' etc.). The agent needs to predict the correct object location which is inside the region. The top-one recall with different distance tolerance is reported in Tab.~\ref{table:area_object_indexing_results}. We notice that VLMaps~\citep{huang23vlmaps} struggles to find the goal in the correct region because VLMaps integrates visual-language features from the encoder fine-tuned on the instance segmentation dataset, improving its segmentation performance on common objects while dropping its ability to recognize more general concepts like regions. In contrast, ConceptFusion integrates pre-trained CLIP features into the map without fine-tuning, enabling it to recognize general concepts including regions, and thus the indexing results are improved.

\begin{table}[h]
  \setlength\tabcolsep{2.5pt}
  \footnotesize\sf\centering
  \begin{tabularx}{\columnwidth}{lccccc}
  \toprule

\multirow{2}{*}{Method}                                            & \multicolumn{4}{c}{Recall@1 (\%)}                                 & \multirow{2}{*}{\begin{tabular}[c]{@{}c@{}}Average\\ min. distance (m)\end{tabular}} \\ \cmidrule{2-5}
                                                                   & \textless{}0.5m & \textless{}1m & \textless{}1.5m & \textless{}2m & \\
\midrule

\begin{tabular}[c]{@{}c@{}}baseline (VLMaps)\end{tabular}        & 5.56            & 7.78          & 13.33           & 17.78         & 8.22 \\ 
\begin{tabular}[c]{@{}c@{}} + ConceptFusion\end{tabular}   & 12.22           & 13.33         & 16.67           & 21.11         & 7.60 \\ 
\begin{tabular}[c]{@{}c@{}} + CLIP sparse (Ours)\end{tabular} & \textbf{15.56}           & \textbf{24.44}         & \textbf{31.11}           & \textbf{35.56}         & \textbf{6.17} \\ 

\midrule
\begin{tabular}[c]{@{}c@{}} + GT region map\end{tabular}   & 37.78           & 44.44         & 55.56           & 61.11         & 2.62 \\ 

  \bottomrule
  \end{tabularx}
\caption{The recall (\%) of area-object cross-modal indexing. }
  \label{table:area_object_indexing_results}

\end{table}

\noindent\textbf{Object-Sound goal indexing.} In this setting, we use object goals to disambiguate sound goals. We collected 119 indexing tasks, each of which consist of a sound category and a nearby object category. Each sound category in a scene can be heard at more than one location, introducing ambiguity to the localization scenario. The recall is reported in Tab.~\ref{table:object_sound_indexing_results}. With the combination of object and audio localization modules, our method largely increases the recall rate for localizing the correct sound goal position in ambiguous scenarios.

\begin{table}[h]
  \setlength\tabcolsep{2.5pt}
  \footnotesize\sf\centering
  \begin{tabular}{lccccc}
  \toprule
\multirow{2}{*}{Method}                                             & \multicolumn{4}{c}{Recall@1 (\%)}                                 & \multirow{2}{*}{\begin{tabular}[l]{@{}c@{}}Average\\ min. distance (m)\end{tabular}} \\ \cmidrule{2-5}
                                                                    & \textless{}0.5m & \textless{}1m & \textless{}1.5m & \textless{}2m &                                                                                      \\ 
\midrule
\begin{tabular}[l]{@{}l@{}}baseline (wav2clip) \end{tabular} & 8.40           & 10.08         & 10.92           & 14.29         & 8.52                                                                                 \\
\begin{tabular}[l]{@{}l@{}}baseline (AudioCLIP) \end{tabular} & 26.05           & 35.29         & 36.97           & 42.01         & 5.04                                                                                 \\
\begin{tabular}[l]{@{}l@{}}VLMaps + wav2clip\end{tabular}         & 24.37           & 30.25         & 33.61           & 38.66         & 6.27                                                                                 \\ 
\begin{tabular}[l]{@{}c@{}}VLMaps + AudioCLIP \\ (Ours)\end{tabular} & \textbf{53.78}           & \textbf{65.55}         & \textbf{67.23 }          & \textbf{70.59}         & \textbf{2.74}                                                                                 \\ 
\bottomrule
\end{tabular}
\caption{The recall (\%) of object-sound cross-modal indexing. }
  \label{table:object_sound_indexing_results}

\end{table}

\noindent\textbf{Visual-Object goal indexing.} In visual-object goal indexing tasks, visual clues are used to resolve ambiguity. Given an object category and an image, our method can localize the correct object near the image position with over 60\% of recall for 0.5 meters distance tolerance, as is shown in Tab.~\ref{table:visual_object_indexing_results}.

\begin{table}[h]
  \setlength\tabcolsep{2.5pt}
  \scriptsize
  \footnotesize\sf\centering
  \begin{tabularx}{\columnwidth}{lccccc}
  \toprule
\multirow{2}{*}{Method}  & \multicolumn{4}{c}{Recall@1 (\%)} & \multirow{2}{*}{\begin{tabular}[c]{@{}c@{}}Average\\ min. distance (m)\end{tabular}} \\ \cmidrule{2-5}& \textless{}0.5m & \textless{}1m & \textless{}1.5m & \textless{}2m & \\ 
\midrule
\begin{tabular}[l]{@{}l@{}}VLMaps w/o \\ visual module\end{tabular}         & 7.55           & 9.43         & 11.32           & 11.94         & 11.22 \\ 
\begin{tabular}[l]{@{}l@{}}VLMaps w/ \\ visual module (Ours) \end{tabular} & \textbf{62.26}    &\textbf{66.67}         & \textbf{70.44}           & \textbf{72.32}         & \textbf{3.11} \\ 

\bottomrule
\end{tabularx}
\caption{The recall (\%) of visual-object cross-modal indexing.}
  \label{table:visual_object_indexing_results}

\end{table}

\begin{figure*}[!htp]
	\centering
	\includegraphics[width=1\textwidth]{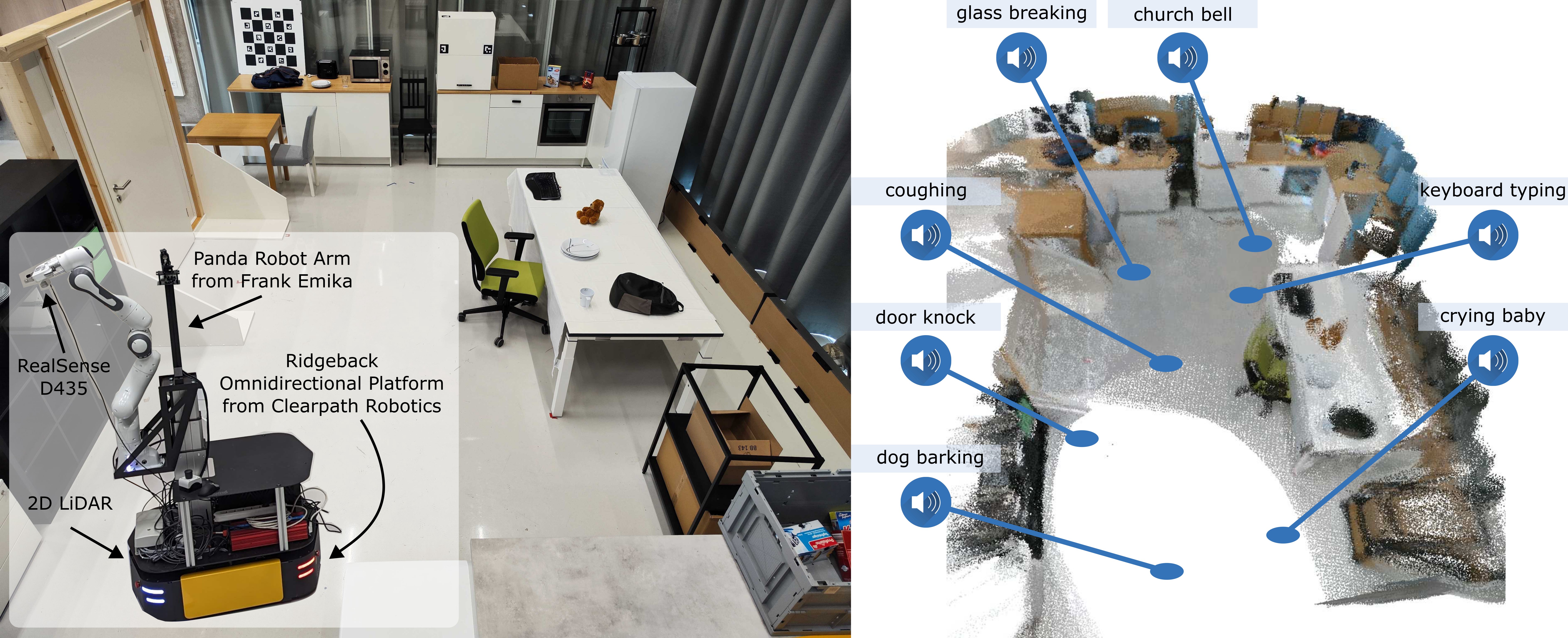}
	\caption{\small Real-world navigation experiments are conducted in a room with multiple ambiguous goals such as tables, chairs, backpacks, and paper boxes (left). The robot setup is also shown in the left image. We leverage dense SLAM techniques to build a 3D reconstruction of the scene from RGB-D camera data into which we anchor features from multiple foundation models (right). We artificially insert sounds with different semantics at locations shown in the image. Different sounds are played when the robot moves to these locations during mapping. Sounds are sampled from the ESC-50 dataset.}
	\label{fig:exp_imbit_2}

\end{figure*}

\subsection{Multimodal Ambiguous Goal Navigation}\label{sec:exp_cross_modal_nav}
In this part, we test our method with ambiguous goal navigation tasks. We collect 119 sequences of tasks. In each task, the agent is required to navigate to an ambiguous sound goal (\ie ``move to the sound of baby crying near the sofa'') and an ambiguous object goal (\ie ``move to the counter near \{image of the kitchen\}'') sequentially. The category of the object near the sound and the image taken near the object are provided. These tasks require the agent to reason across different modalities to accurately localize the target. We consider two single-modality baselines: VLMaps~\citep{huang23vlmaps} and AudioCLIP~\citep{guzhov2022audioclip} and one multimodal baseline. The multimodal baseline uses VLMaps as the object localization module, wav2clip~\citep{wu2022wav2clip} as the audio localization module and the same visual localization module as our method. The results are shown in Tab.~\ref{table:cross_modal_nav}. We observe that AVLMaps navigates to cross-modal goals with a 24.2\% higher success rate to ambiguous sound goals and with a 2.1\% higher success rate to ambiguous object goals compared to the alternative multimodal baseline.

\begin{table}[h]
  \setlength\tabcolsep{2.5pt}
  \footnotesize\sf\centering
  \begin{tabular}{lccccc}
  \toprule
\multirow{2}[1]{*}{Tasks}  & \multicolumn{2}{l}{No. Subgoals in a Row} & Sound  & Object    \\                
                    \cmidrule(lr){2-3}
               &   1             & 2                        &  Goals   & Goals \\
\midrule
VLMaps        &    -            &       -                  &    -      &   27.1       \\
AudioCLIP      &    -            &       -           &    16.9       &   -    \\
VLMaps + wav2clip & 22.0    & 12.7        & 22.0 & 53.4     \\
\begin{tabular}[l]{@{}c@{}}VLMaps + AudioCLIP (Ours)\end{tabular} & \textbf{46.2}    & \textbf{28.6}        & \textbf{46.2} & \textbf{55.5}     \\

  \bottomrule
  \end{tabular}
  \caption{The success rate (\%) of multimodal ambiguous goal navigation with AVLMaps. The agent is required to navigate to one ambiguous sound goal and one ambiguous object goal sequentially.}
  \label{table:cross_modal_nav}

\end{table}

\begin{table}[h]

  \setlength\tabcolsep{4pt}
  \footnotesize\sf\centering
  \begin{tabular}{llcccccc}
  \toprule
\multirow{2}[1]{*}{ALM} & \multirow{2}[1]{*}{VLM}  & \multicolumn{4}{l}{No. Subgoals in a Row} & Sound  & Object       \\                
                    \cmidrule(lr){3-6}
                          &                   &  1              & 2               & 3             & 4               &                &      \\
\midrule
AudioCLIP                 &     LSeg          & 71.0            & 59.0            & 29.0          &   17.0          & 71.0           & 41.8 \\
CLAP                      &     LSeg          & \textbf{81.0}   & \textbf{69.0}   & 36.5          &   19.0          & \textbf{81.0}  & 52.0 \\
CLAP                      &     SAM+CLIP      & 80.0            & 68.5            & 29.5          &   12.0          & 80.0           & 40.5 \\
CLAP                      &     OVSeg         & \textbf{81.0}   & \textbf{69.0}   & \textbf{37.0} &   \textbf{21.0} & \textbf{81.0}  & \textbf{53.5}\\

  \bottomrule
  \end{tabular}
  \caption{The success rate (\%) of multimodal goal navigation with different foundation models. The agent is required to navigate to one sound goal, one visual goal, and two object goals in sequence. The right-most three columns indicate the independent success rate of navigating to specific types of goals.}
  \label{table:scale}

\end{table}

\subsection{Scaling Experiment}\label{sec:exp_scale}

Since AVLMaps is highly modular, each module can be upgraded with advanced foundation models that generate similar audio-language features or visual-language features. In this section, we explore whether AVLMaps can evolve with the advancement in foundation model research. We follow similar settings of multimodal goal navigation as in Sec.~\ref{sec:exp_multi_modal_goal_nav} to test the AVLMaps modules supported by different foundation models. In this experiment, the robot needs to navigate to a sound goal, a visual goal, and two object goals in a sequence and we report the in-a-row navigation success rate as well as the success rate for each type of goal. In this experiment, we mainly focus on analyzing how the performance scales with improved Audio Language Models and Vision Language Models. We fixed the visual localization module as NetVLAD and SuperGLUE.
\\
\\
\noindent\textbf{Audio Language Models.} We compare the downstream performance of using AudioCLIP~\citep{guzhov2022audioclip} with respect to a more recent CLAP~\citep{elizalde2023clap} model. 
\\
\\
\noindent\textbf{Vision Language Models.} In order to leverage pretrained Vision Language Models, we require the encoder to generate pixel-wise features in CLIP embedding space. We compared LSeg~\citep{li2021language}, the VLM used in VLMaps~\citep{huang23vlmaps}, with OVSeg~\citep{liang2023ovseg} and a method that uses SAM~\citep{kirillov2023sam} and CLIP~\citep{radford2021learning} introduced in HOV-SG~\citep{werby2024hovsg}. The method in HOV-SG first leverage SAM~\citep{kirillov2023sam} to generate class-agnostic masks. Each region cropped with a mask and its zero-background version are encoded with a CLIP image encoder and their embeddings are summed with weights. The resulting embedding is assigned to all pixels in the masked region.
\\
\\
\noindent\textbf{Results.} We report the results in Table~\ref{table:scale}. The first two rows compare the success rates of different Audio Language Models (ALMs) using the same VLM. We observe a significant 10\% improvement in navigating to sound goals when replacing AudioCLIP with CLAP. This improvement is largely due to CLAP’s more extensive pretraining on larger, more diverse datasets with advanced augmentation techniques~\citep{elizalde2023clap}. From the second to the final rows, we compared three VLMs that generate dense visual-language features but fixed CLAP as the ALM. However, recent VLMs did not offer clear benefits. In fact, the SAM+CLIP combination significantly reduced performance. Upon closer inspection, we found that while SAM+CLIP performed well in previous work~\citep{werby2024hovsg}, it requires extensive hyper-parameter tuning to adapt to different scenes. The quality of SAM’s masks is highly sensitive to its parameters, and CLIP struggles to interpret masked regions, as highlighted by OVSeg~\citep{liang2023ovseg}. OVSeg addressed this by training learnable mask prompts, and improving 2D semantic segmentation through ensemble techniques. However, these methods are designed to enhance segmentation performance, not visual-language feature generation. Like LSeg, OVSeg is fine-tuned on similar datasets, offering limited improvements in dense pixel-level visual-language features. As a result, AVLMaps’ object localization module, which relies on these features, sees minimal benefit from OVSeg. However, we believe that pixel-level visual language features can be improved with access to more high-quality segmentation datasets in the future, boosting the performance of the object localization module in AVLMaps.

\subsection{Real World Experiment for Mobile Robot}
\label{sec:exp_real_world}

To answer the question of how AVLMaps applies to real-world environments and benefits multimodal navigation, we designed a mobile navigation experiment in which the robot must locate sound, image, and object-based goals within an environment containing duplicate objects from certain categories such as chairs, backpacks, shelves and so on. This setup demonstrates how AVLMaps enables the robot to retrieve multimodal concepts and disambiguate goals by leveraging information from additional modalities.
\\
\\
\noindent\textbf{Robot Setup.} In the real-world experiment setting, we use a mobile robot equipped with a Ridgeback omnidirectional platform from Clearpath Robotics as the mobile base, and a Panda manipulator from Franka Emika. We mount a RealSense D435 RGB-D camera at the gripper of the Panda manipulator. During the mapping, we run a LiDAR localizer to provide the odometry for the robot base and derive the camera pose through the forward kinematics of the robot arm.

\noindent\textbf{Environment Setup.} We choose a room with multiple ambiguous goals such as tables, chairs, paper boxes, counters, and backpacks, which are shown on the left in Fig.~\ref{fig:exp_imbit_2}. We control the robot in this environment and record RGB-D video. Then we artificially add sounds to the RGB-D video when the robot moves to certain locations. The sound locations are shown on the right in Fig.~\ref{fig:exp_imbit_2}.

\begin{figure}[!t]
	\centering
	\includegraphics[width=1\linewidth]{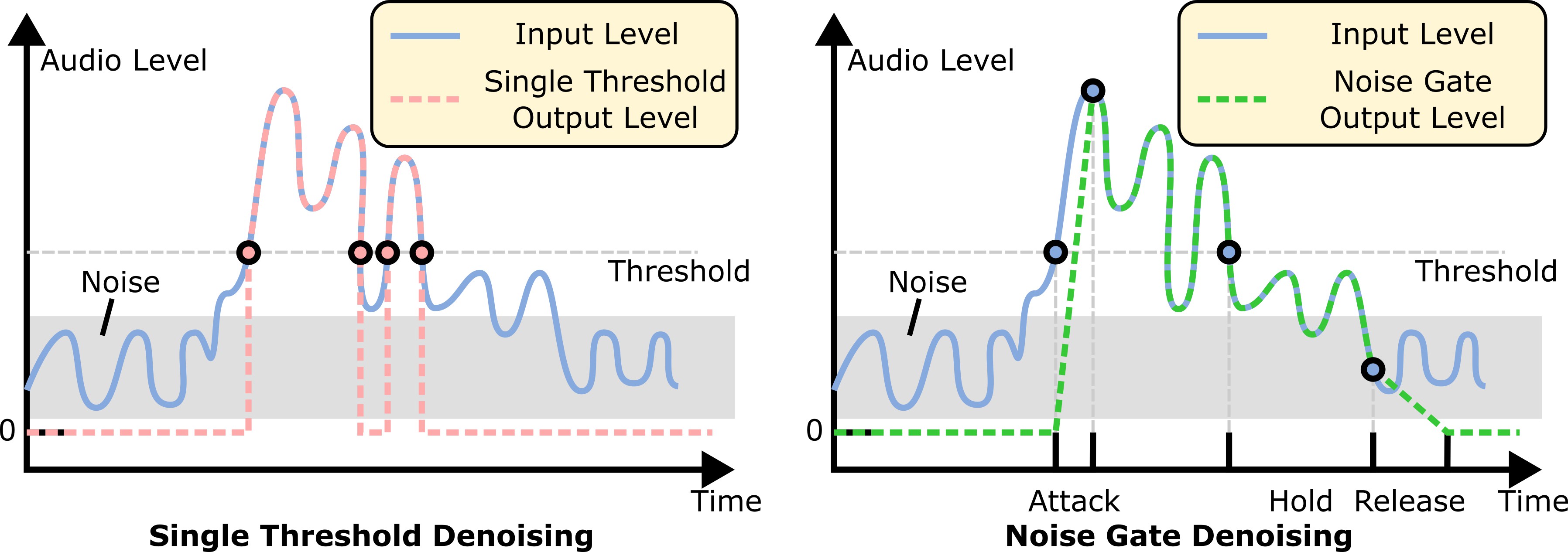}
	\caption{\small Audio denoising for real-world experiment. We pre-process the audio by applying a noise gate. Simply applying a threshold for filtering out low-level noise might lead to frequent fluctuation of the audio level, leading to fragmentation of the target audio (see the left). A noise gate contains ``attack'', ``hold'', and ``release'' phases, which introduce smooth transitions and prevent cutting off short audio signals (see the right).}
	\label{fig:denoising}

\end{figure}

\noindent\textbf{Audio Recording and Denoising.} To make the experiment more realistic, instead of simply adding clean audio to the video, we recorded environmental noise using a microphone during the robot’s exploration. We then overlaid semantic soundtracks from ESC-50 onto the recorded environmental noise at specific locations. This approach provides an approximation of real-world audio conditions with background noise. We also experimented with playing sounds through a speaker in the environment and recording both the environmental noise and semantic audio simultaneously. However, the mobile robot produced significant vibration noise while moving, which masked the played sounds and rendered them largely inaudible. As a result, we approximated the noisy semantic audio by mixing the semantic soundtracks with the recorded environmental noise.

Since current audio language models are sensitive to noise in input audio, we apply a noise filtering algorithm as a pre-processing step to the audio before sending them to the audio localization module. In our case, we use a noise gate, as shown in Fig.~\ref{fig:denoising}. A noise gate applies thresholding with smooth fade-in and fade-out transitions. Instead of simply zeroing out values below the threshold, it includes three phases: attack, hold, and release. When the input audio level exceeds the threshold, the output ramps up linearly from zero to the input level over a period defined by the attack time. Once the input drops below the threshold, the output stays at the input level for the duration of the hold time. If the input remains below the threshold after the hold period, the output level decreases linearly to zero over the release time. This approach effectively reduces environmental noise. We then apply silence segmentation to further extract meaningful audio clips. In our setup, the noise gate threshold is set to –10 dB ($\text{dB} = 20 * \log10(\text{amplitude})$), with an attack time of 250 ms, hold time of 1000 ms, and release time of 170 ms. The silence segmentation threshold is set to 0.1. As noise cancellation is not the primary focus of this work, these parameters were selected based on empirical experience. In the future, a more robust approach would involve adding random real-world noise during audio language model pretraining to improve noise tolerance.
\\
\\
 \noindent\textbf{Map Building and Navigation.} After collecting the data, we run the AVLMaps mapping offline. For navigation tasks, we provide the AVLMaps and the language instruction as input. The robot parses the instruction (Sec.~\ref{subsec_multimodal_nav_from_language}) and executes the generated Python code for goal indexing and planning. We use the ROS navigation package~\citep{quigley2009ros} for global and local planning. We pre-process sound inputs with background noise subtraction to avoid including noise from the robot operation.

 \begin{figure*}[htb]
	\centering
	\includegraphics[width=1\textwidth]{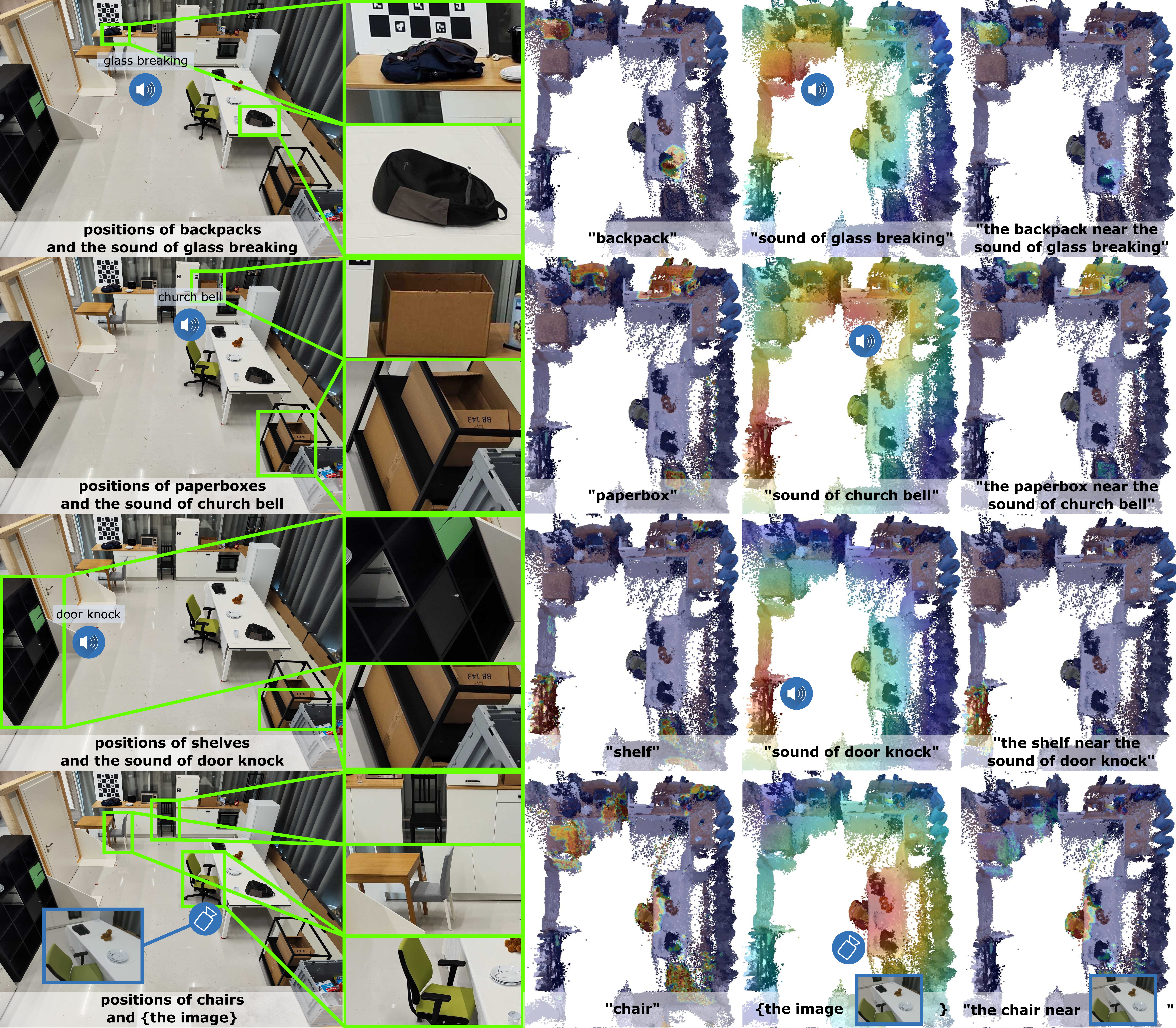}
	\caption{\small Visualization of example heatmaps in AVLMaps for multimodal goal reasoning for ambiguous object goals. The first column shows the positions of ambiguous objects (green bounding boxes) and the location of a sound (the icon of a speaker) or an image (the icon of a camera), while the second column shows the close-up view of ambiguous objects in the scene. The third column shows the predicted 3D heatmap for the object. The fourth column shows the heatmap for the extra modality, and the fifth column shows the fused heatmap after cross-modal reasoning. Sounds are artificially inserted into the scene for benchmarking and evaluation. The locations of sounds are not sound-source locations but the places where the sounds were heard. The heatmap is shown in the JET color scheme (red means the highest score and blue means the lowest score).}
	\label{fig:real_object_heatmap}

\end{figure*}

 \begin{figure*}[htb]
	\centering
	\includegraphics[width=1\textwidth]{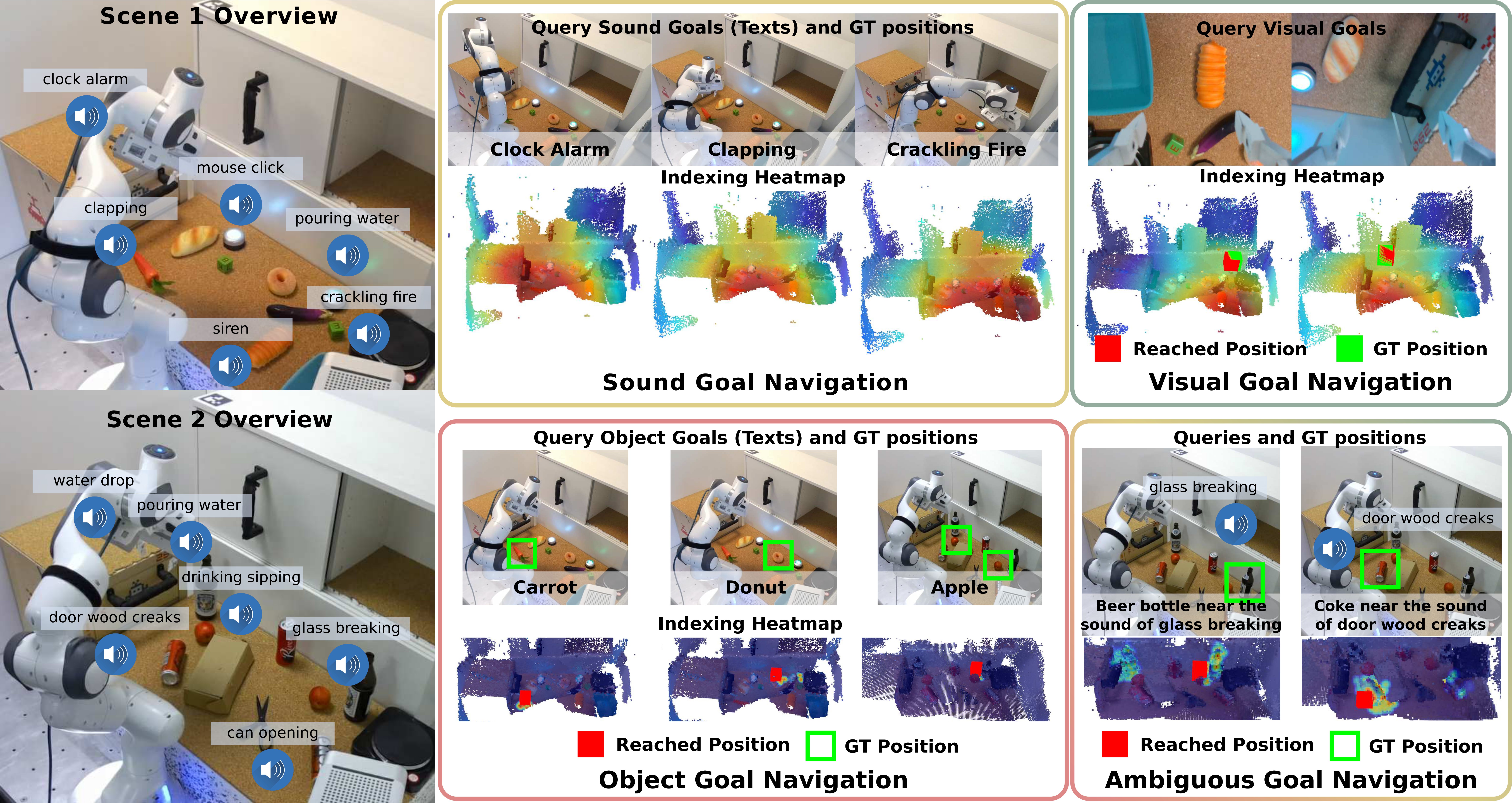}
	\caption{\small Visualization of tabletop goal-reaching experiments. We set up two tabletop scenes and inserted sounds to different locations in the observation data (left). We show the indexing heatmap results of sound goals, visual goals, object goals, and ambiguous goal reaching results on the right. In addition, for visual goals, object goals, and ambiguous goals, we show the ground truth target locations (green cubes or green boxes) as well as the reached positions (red cubes). The heatmap is shown in the JET color scheme (red means the highest score and blue means the lowest score).}
	\label{fig:real_tabletop_heatmap}

\end{figure*}

\noindent\textbf{Multimodal Spatial Goal Reasoning and Navigation with Natural Language.}  We design 20 language-based multimodal navigation tasks, asking the robot to navigate to sounds, images, and objects. We report an overall success rate of 50\%. We also design an evaluation consisting of ten multimodal spatial goals. The agent needs to reason across object, sound, image and spatial concepts. An example is ``navigate in between the backpack near the sound of glass breaking and \{the image of a fridge\}''. In the end, six out of ten tasks were successfully finished. We show in Fig.~\ref{fig:real_object_heatmap} the process of resolving ambiguities in the scene. There are different ambiguous objects in the scenes including paper boxes, backpacks, shelves, tables, chairs, and plates. The first and the second columns in Fig.~\ref{fig:real_object_heatmap} show the ground truth positions of the target objects and sounds. The third and fourth columns show that AVLMaps can accurately localize objects, sounds, and visual goals in the form of 3D heatmaps. The final column shows that our method can correctly narrow down targets in spite of object ambiguities. We can observe from the figure that AVLMaps can accurately localize ambiguous concept with language, audio and image. We observe that the failures come from the composition of the imperfection of different modules. For example, the object localization module (\eg VLMaps) fails to recognize rare objects like various toys. It also mistakes some shelves for chairs. Similar failures happen in audio localization module. In the second row and the fourth column in Fig.~\ref{fig:real_object_heatmap}, the church bell sound should be at the top-right corner but the module also gives high score for the sound heard at bottom-left.


\subsection{Real World Experiment for Table-Top Goal Reaching}\label{sec:exp_table_top}

In the previous section, we demonstrated how AVLMaps empower a robot to interpret multimodal goals in room-scale mobile navigation tasks. Here, we extend our investigation to assess how AVLMaps benefit a fixed-based manipulator in real-world table-top tasks, which require a more detailed semantic understanding of the scene. In this setup, the robot manipulator must approach multimodal goals with a stricter tolerance for error (within 10 cm). Additionally, we explore AVLMaps' potential for application across robots with varied embodiments.

\noindent\textbf{Robot setup.} We set up a Panda robot arm from Franka Emika on a table and mount a FRAMOS D345e industrial RGB-D camera at the gripper of the manipulator. Before the experiment, we calibrate the extrinsic matrix from the end effector coordinate frame to the camera frame. We use an HTC Vive VR controller to teleoperate the robot end effector to collect observation data.

\noindent\textbf{Experiment setup.} We set up two tabletop scenes where different objects are lying on the table at random locations. In one scene, we deliberately place duplicate objects on the table to simulate the ambiguity of objects. Subsequently, we teleoperate the robot end effector with a VR controller and collect observations including the RGB, the depth, and the robot end effector poses relative to the robot base coordinate frame. The recording frequency is 30 Hz. Later, we derive the camera poses relative to the base coordinate frame using the end effector pose and the extrinsic matrix obtained during the calibration. For each scene, we first collect a sequence of data for generating maps with the object localization module and the visual localization module. We then control the robot to different areas on the table and collect an episode of data in each region, to which we later insert a random segment of audio sampled from ESC-50 dataset in a similar way as the mobile robot experiment setting. These observation data augmented with audio are used to generate the audio map with the audio localization module. Thanks to the insights we obtained from the scaling experiments in Sec.~\ref{sec:exp_scale}, we use CLAP~\citep{elizalde2023clap} and OVSeg~\citep{liang2023ovseg} as our foundation models for the audio localization module and the object localization module. For the visual localization module, we still use the NetVLAD~\citep{arandjelovic2016netvlad} and SuperGLUE~\citep{sarlin2020superglue} scheme as in Sec.~\ref{subsec_building_avlmaps}.

\noindent\textbf{Tabletop manipulator goal reaching.} We randomly moved the robot arm and collected a sequence of RGB images as the visual goals used for querying the AVLMaps created earlier. During inference, we prompted the robot with randomly selected visual goals (sampled from the collected images), sound categories (matching the inserted audio types), and object categories (language descriptions of objects on the table), instructing it to approach the target region. If the final position of the robot's end effector was within 10 cm of the ground truth, the trial was considered successful. We tested 20 visual goals, twelve sound goals, and 13 object goals. The robot successfully reached 100\% of the visual and sound goals, and nine out of 13 object goals. Additionally, we tested ten ambiguous goals, such as ``the light near the sound of clock alarm'' or ``the apple near the image \{path/to/image\}'' and the robot successfully approached nine out of ten. In summary, AVLMaps demonstrated excellent performance in a tabletop setting, successfully navigating to multimodal goals, including ambiguous ones. The results are shown in Fig.~\ref{fig:real_tabletop_heatmap}.

\section{Limitations and Discussions}
\label{sec:limitations_and_discussions}
While our multimodal spatial language maps approach is versatile in terms of navigating to various spatially grounded concepts and is effective in the disambiguation of duplicate goals with extra information, it does have certain limitations. In this section, we thoroughly discuss the major drawbacks of our method, along with potential directions for extension and improvement. Through this analysis, we aim to offer insights to the community and inspire future research.

\noindent\textbf{Transient Sound Reaction.} One of the main challenges with our AVLMaps is that sound is inherently transient, while the map relies on pre-exploration and offline mapping to embed the information for navigation tasks. Even if sounds are associated with specific locations during the exploration phase, those sounds might disappear or shift to new locations by the time of the inference and navigation. Therefore, AVLMaps struggles to support reactive navigation towards transient sound goals. Previous work on vision and audio navigation~\citep{chen2020soundspaces,younes2023catch,gan2020look} has focused on enabling robots to respond to transient sounds using biaural and visual observations. However, these methods are limited to transient goals and have been tested only in simulated environments. We argue that both transient sounds and previously heard sounds are essential for intelligent navigation. Transient sounds serve as important cues for immediate action, while past sounds provide valuable references for narrowing down possible targets based on prior experiences. For example, when instructed to go to ``the caf\'e where you heard the song'', we can recall the specific caf\'e and adjust our navigation accordingly. A robot must be able to understand and navigate toward both kinds of sound goals to achieve human-level intelligence in sound-based navigation. However, no current system integrates these dual capabilities. We believe this gap presents an exciting opportunity for future research.

\noindent\textbf{Sound Localization.} While AVLMaps manages to comprehend the semantic meaning of sounds, it struggles to localize the sound sources. In this paper, the audio localization module assumes monaural audio input, lacking in the ability to utilize binaural audio to localize sound sources with triangulation like humans. More specifically, a sound can be heard in all locations inside a room, but we only associate its features with the robot's current location. Encouragingly, several concurrent works are actively addressing the sounds source localization, including efforts to learn the real-world acoustic sound field~\citep{chen2024real_acoustic_fields}, reconstruct the acoustic properties of environments~\citep{wang2024hearing_anything}, and develop more acoustically realistic simulation environments~\citep{chen2022soundspaces20}.

\noindent\textbf{Dynamic Scenes.} Another challenge faced by our multimodal spatial language maps (both VLMaps and AVLMaps) is their inability to handle dynamic scenes. One form of dynamics involves in-view dynamics, such as walking humans and moving articulated objects during exploration. These dynamic entities easily corrupt the object map, leaving behind point artifacts that trace their movement trajectories. The semantic features of the moving objects may be erroneously associated with these artifact points, leading to inaccurate representations of their true locations. Another form of dynamics involves long-term dynamics, such as object relocations between the exploration and inference phases. These relocations can invalidate the pre-built map, necessitating an updating mechanism to ensure accurate navigation. Currently, our multimodal spatial language maps lack mechanisms to address both types of dynamics. To explore potential solutions, we investigated approaches to mitigating the impact of dynamics. Prior works have addressed in-view dynamics by removing or tracking certain classes of semantic masks during mapping~\citep{xu2019mid,runz2018maskfusion}. For long-term dynamics, recent research has proposed learning-based object association methods to update the locations of relocated objects in the map~\citep{yan2025dynamic_open_vocab_scene_graph,huang2025bye}. These solutions could be integrated into our mapping pipeline to enhance its robustness.

\noindent\textbf{Extension to More Modalities.} In this paper, we have demonstrated the feasibility of integrating information from multiple modalities into a unified map representation. More broadly, our method provides a framework that can be extended to additional modalities, such as odor, temperature, magnetic fields, infrared imagery, and point clouds. In our implementation, we selected audio, language, and vision because they closely mirror human perceptual capabilities. Viewed from a broader perspective, AVLMaps can be regarded as a case study in building a multimodal spatial memory system, with inherent flexibility to incorporate more modalities. Extending the system to a new modality X involves three steps: (i) implementing the ``X localization module'' which includes mapping (embedding and storing the data into a representation) and retrieval (generating a heatmap based on the query data and the map) functions for the new modality, (ii) implementing the \texttt{get\_major\_map(X\_input=None)} and \texttt{get\_map(X\_input=None)} to accept a new modality's data as input and return 3D heatmaps with high and low decay rates as in Sec.~\ref{subsec_multimodal_nav_from_language}, and (iii) implementing a context prompt as in Fig.~\ref{fig:full_prompt} for the new modality, including an instruction involving the new modality, and the expected generated code. Although AVLMaps only focuses on audio, language, and vision, its simplicity, flexibility, and scalability open up promising directions for future research.

\section{Conclusion}

In this paper, we introduced multimodal spatial language maps, a unified mapping framework that is spatial, multimodal, reusable, and extensible. We first introduced a visual language map representation, VLMaps, that enables robots to navigate to long-horizon spatial goals in a zero-shot manner. By defining the categories where the robot can and can not traverse, our map representation can adaptively generate obstacle maps for different embodiments, allowing for efficient path planning. Subsequently, we further extend our visual-language maps to a multimodal version, AVLMaps, which is a unified 3D spatial map representation for storing cross-modal information from audio, visual, and language cues. AVLMaps retain the spatial and reusable properties of VLMaps while enabling robots to reason over multimodal cues to disambiguate goals using large language models. Experiments in both simulated and real-world environments with different robotic embodiments demonstrate that our multimodal spatial language maps enable zero-shot spatial and multimodal goal navigation, significantly outperforming baselines in navigation success rate and landmark indexing accuracy, especially in scenarios with ambiguous goals. Moreover, extensive experimentation reveals that the performance of multimodal navigation and manipulation tasks scales with the capabilities of the underlying foundation models. At the end of the paper, we also present an in-depth discussion of the limitations and potential directions for future work, to inspire further research in multimodal spatial reasoning for robotics.

\theendnotes
\bibliographystyle{SageH}

\bibliography{references}

\end{document}